\documentclass[lettersize,journal]{IEEEtran}
\usepackage{amsmath,amsfonts}
\usepackage{algorithmic}
\usepackage{algorithm}
\usepackage{array}
\usepackage[caption=false,font=normalsize,labelfont=sf,textfont=sf]{subfig}
\usepackage{textcomp}
\usepackage{stfloats}
\usepackage{url}
\usepackage{verbatim}
\usepackage{graphicx}
\usepackage{cite}
\usepackage[colorlinks, linkcolor=blue, citecolor=blue, urlcolor=blue]{hyperref}
\usepackage{algorithm}
\usepackage{algorithmic}
\usepackage{amsmath}
\usepackage{amsfonts}
\usepackage{multirow} % 引入multirow宏包  
\usepackage{booktabs}
\usepackage{siunitx}
\usepackage{color} 
\usepackage{amssymb}
\usepackage{float}  % 导言区加入
\newcommand{\concat}{\mathbin{\circledast}}
\begin{document}

\title{Bidirectional Feature-aligned Motion Transformation for Efficient Dynamic Point Cloud Compression}

\author{Xuan Deng, Xingtao Wang, Xiandong Meng, Longguang Wang, Tiange Zhang,\\
Xiaopeng Fan,~\IEEEmembership{Senior Member, IEEE}, and Debin Zhao
        % <-this % stops a space
\thanks{Xuan Deng, Xiaopeng Fan and Debin Zhao are with the School of Computer Science and Technology, Harbin Institute of Technology, Shenzhen 518055, China, and also with the Peng Cheng Laboratory, Shenzhen 519055, China (e-mail: 24b951089@stu.hit.edu.cn; fxp@hit.edu.cn; dbzhao@hit.edu.cn).}
\thanks{Xingtao Wang is with the School of Computer Science and Technology, Harbin Institute of Technology, Harbin, China (e-mail: xtwang@hit.edu.cn)}
\thanks{Xiandong Meng and Tiange Zhang are with the Peng Cheng Laboratory, Shenzhen 519055, China (e-mail: mengxd@pcl.ac.cn; zhangtg@pcl.ac.cn)}
\thanks{Longguang Wang is with the Sun Yat-sen University, Guangzhou, China (e-mail: wanglg9@mail.sysu.edu.cn)}
}

% \markboth{Journal of \LaTeX\ Class Files,~Vol.~14, No.~8, August~2025}%


\maketitle

\begin{abstract}
Efficient dynamic point cloud compression (DPCC) critically depends on accurate motion estimation and compensation. However, the inherently irregular structure and substantial local variations of point clouds make this task highly challenging. Existing approaches typically rely on explicit motion estimation, whose encoded motion vectors often fail to capture complex dynamics and inadequately exploit temporal correlations.
To address these limitations, we propose a \textbf{Bidirectional Feature-aligned Motion Transformation (Bi-FMT)} framework that implicitly models motion in the feature space. Bi-FMT aligns features across both past and future frames to produce temporally consistent latent representations, which serve as predictive context in a conditional coding pipeline, forming a unified ``Motion + Conditional'' representation.
Built upon this bidirectional feature alignment, we introduce a Cross-Transformer Refinement module (CTR) at the decoder side to adaptively refine locally aligned features. By modeling cross-frame dependencies with vector attention, CRT enhances local consistency and restores fine-grained spatial details that are often lost during motion alignment.
Moreover, we design a \textbf{Random Access (RA)} reference strategy that treats the bidirectionally aligned features as conditional context, enabling frame-level parallel compression and eliminating the sequential encoding.
Extensive experiments demonstrate that Bi-FMT surpasses \textbf{D-DPCC} and \textbf{AdaDPCC} in both compression efficiency and runtime, achieving BD-Rate reductions of \textbf{20\% (D1)} and \textbf{9.4\% (D1)}, respectively.
\end{abstract}


\section{Introduction}
\IEEEPARstart{A} Dynamic point cloud (DPC) is a sequence of point cloud frames, each consisting of unordered points sparsely distributed in 3D space~\cite{xu2018introduction}. As a promising 3D data representation, DPC has found broad applications in immersive media~\cite{gao2023point}, volumetric capture, and AR/VR~\cite{li2019advanced,xie2024roi}. Nevertheless, its massive data volume poses significant challenges for storage and transmission, limiting widespread adoption.

% The Moving Picture Experts Group (MPEG) has approved two point cloud compression standards~\cite{schwarz2018emerging,graziosi2020overview}: Geometry-based Point Cloud Compression (G-PCC)~\cite{li2024mpeg} and Video-based Point Cloud Compression (V-PCC). G-PCC includes octree-geometry coding as a generic geometry coding tool and a predictive geometry coding (tree-based) tool which is more targeted toward LiDAR-based point clouds. G-PCC is still developing
% triangle meshes or triangle soup (trisoup) based methods to approximate the surface of the 3D model. V-PCC on the other hand encodes dynamic point clouds by projecting 3D points onto a 2D plane and then uses video codecs, e.g., High Efffciency Video Coding (HEVC), to encode each frame over time. In addition to the standardized frameworks by the Moving Picture Experts Group (MPEG), a distinct category of point cloud compression methods performs inter-prediction using hand-crafted temporal context selection algorithms.
% For instance, the method in~\cite{santos2021block} employs various point cloud partitioning and matching strategies to identify correspondences between regions. Similarly, a motion-compensated approach for encoding dynamic voxelized point clouds at low bitrates was introduced in~\cite{de2017motion}. Other related techniques leverage projection-based algorithms to manage temporal redundancies~\cite{zhu2020view}.

The Moving Picture Experts Group (MPEG) has approved two point cloud compression standards~\cite{schwarz2018emerging,graziosi2020overview}: Geometry-based Point Cloud Compression (G-PCC)~\cite{li2024mpeg}, which uses octree and predictive geometry coding for LiDAR-based point clouds and triangle-based surface approximation, and Video-based Point Cloud Compression (V-PCC)~\cite{graziosi2020overview}, which projects 3D points onto 2D planes and encodes them with video codecs such as HEVC~\cite{sullivan2012overview}. 
Beyond these standards, other methods perform inter-frame prediction using hand-crafted temporal context, including region partitioning and matching~\cite{santos2021block}, motion-compensated voxel encoding~\cite{de2017motion}, or projection-based techniques to reduce temporal redundancy~\cite{zhu2020view,cai2024distortion}. Inspired by advances in neural image and video compression, dynamic point cloud compression (DPCC) aims to eliminate spatial and temporal redundancies. Spatial redundancy is typically managed by using Variational Autoencoders (VAEs) to encode features into a compact latent space \cite{huang2024temporal,zhang2024content}. For temporal redundancy, current approaches rely on motion estimation and KNN-based motion compensation in the latent domain \cite{fan2022d,xia2023learning}.
Methods like D-DPCC \cite{fan2022d} and its successors \cite{xia2023learning,jiang2023end,jiang2025mp} primarily use explicit motion estimation (e.g., KNN-based 3DAWI) to predict frames and compress the residual. However, non-uniform geometry and scale variation make explicit motion estimation highly challenging. To address this, some works resort to implicit motion capture via target convolution \cite{akhtar2024inter} or Temporal Information Embedding (TIE) \cite{liu2024encoding}.
Crucially, all these mainstream techniques are built upon the fundamental "Motion + Residual" architecture. 

\textcolor{black}{Recently, several dynamic point cloud compression frameworks based on conditional coding~\cite{zhang2025adadpcc,2025A} have demonstrated superior performance over traditional ``Motion + Residual'' approaches, which typically depend on explicit motion estimation whose encoded motion vectors struggle to represent complex dynamics and insufficiently exploit temporal correlations.}
To enable more robust motion modeling, we reformulate conventional motion estimation and compensation as a spatiotemporal alignment task. Specifically, we introduce a point-based Bidirectional Feature-aligned Motion Transformation (Bi-FMT) framework to implicitly model dynamic variations. Bi-FMT achieves higher adaptability to local deformations than explicit motion models by aligning features across both past and future frames. 
To further enhance the locally aligned features, a Cross-Transformer refinement (CTR) module is incorporated at the decoding stage to adaptively aggregate cross-frame information, thereby improving local consistency and preserving fine-grained spatial details. 
Bi-FMT produces temporally consistent latent representations, which serve as predictive context in a conditional coding pipeline, thus forming a unified ``Motion + Conditional'' representation. 
Furthermore, we integrate a Random Access (RA) reference strategy that utilizes the bidirectionally aligned features as context for prediction, enabling a non-sequential hierarchical structure that replaces conventional frame-by-frame encoding.
The contributions of our work are as follows:

\begin{itemize}
    \item The Bidirectional Feature-aligned Motion Transformation (Bi-FMT) module reformulates motion estimation and compensation as a spatiotemporal alignment task. It implicitly aligns bidirectional frame features in the latent space, enabling robust and flexible motion representation without relying on explicit motion estimation.

    \item To enhance locally aligned features, a Cross-Transformer Refinement (CTR) module is designed to adaptively aggregate cross-frame information, improving local consistency and preserving fine-grained geometric details for non-rigid motion.

    \item Built upon a hierarchical non-sequential reference structure, the Random Access (RA) mode dynamically selects forward and backward references, supporting frame-level parallel compression and improving overall coding efficiency.

    \item Extensive experiments demonstrate that our framework achieves substantial gains in both encoding efficiency and compression performance compared with representative baselines such as D-DPCC and AdaDPCC, achieving notable BD-Rate reductions on multiple distortion metrics.
\end{itemize}

\section{Related Work}

\subsection{Dynamic Point Cloud Compression}

Dynamic point cloud compression has been widely explored, with methods broadly categorized into rule-based and learning-based approaches. Rule-based methods, such as the MPEG V-PCC standard~\cite{schwarz2018emerging}, project 3D point clouds onto 2D planes and compress them using traditional video codecs~\cite{li2019advanced}. Other techniques, including PatchVVC~\cite{chen2023patchvvc} and the framework by Mekuria et al.~\cite{mekuria2016design}, reduce inter-frame redundancy in 3D space through block or patch matching. However, these methods rely on handcrafted motion estimation, which often struggles to capture localized movements in large-scale, dense point clouds, especially under non-uniform distributions or scale variations.

Learning-based methods have achieved remarkable success in both static and dynamic point cloud compression~\cite{wang2022sparse, zhang2023yoga, wang2024versatile, wang2021multiscale}. In static point cloud compression~\cite{fu2022octattention,cui2023octformer,nguyen2023lossless,gao2023point,guo2024tsc}, most approaches primarily focus on eliminating spatial redundancy within coordinate distributions.
For dynamic point cloud compression, temporal redundancy is additionally exploited through inter-frame prediction. A common line of research conducts motion estimation and compensation between the current frame and a single forward reference, followed by residual encoding. Representative examples include Unicorn~\cite{wang2024versatile} that integrates multiscale motion compensation with context modeling, D-DPCC~\cite{fan2022d} that learns motion estimation in an end-to-end manner, and the coarse-to-fine refinement strategy of Xia et al.~\cite{xia2023learning}. Jiang et al.~\cite{jiang2023end} further introduce multiscale motion refinement by conditioning fine motion on coarse priors, while MP-VPC~\cite{jiang2025mp} adopts a feature proxy module to locate representative anchor points for motion modeling.
Other approaches implicitly infer temporal correspondence in latent space without regressing explicit motion vectors. Akhtar et al.~\cite{akhtar2024inter} compensate the current frame’s latent representation using reference neighbors via target convolution, whereas AuxGR~\cite{liu2024encoding} expands the temporal modeling capability by combining KNN and self-attention. Although computationally efficient, these methods also rely on unidirectional prediction and provide limited improvements in compression performance.

% In practice, existing inter-frame prediction strategies still face challenges in accurately modeling subtle and spatially localized geometric variations commonly observed in real-world dynamic scenes. In addition, their strictly sequential processing pipeline, where motion estimation, motion compensation, and residual encoding are executed in a fixed order, limits the flexibility of temporal context exploitation.
% In this work, motion modeling is reformulated as a spatiotemporal alignment problem. We propose a Bidirectional Feature-aligned Motion Transformation (Bi-FMT) module that implicitly aligns frame features from both past and future directions without relying on explicit motion estimation. In addition, a non-sequential hierarchical encoding paradigm is introduced to select both past and future reference frames, enabling more adaptive temporal context utilization and improving compression efficiency.
In practice, existing inter-frame prediction strategies still face challenges in accurately modeling subtle and spatially localized geometric variations commonly observed in real-world dynamic scenes. In addition, their strictly sequential processing pipeline, where motion estimation, motion compensation, and residual encoding are executed in a fixed order, limits the flexibility of temporal context exploitation.
In this work, motion modeling is reformulated as a spatiotemporal alignment problem. We propose a Bidirectional Feature-aligned Motion Transformation (Bi-FMT) module that implicitly aligns frame features from both past and future directions without relying on explicit motion estimation. Bi-FMT produces temporally consistent latent representations, which serve as predictive context in a conditional coding pipeline, thus forming a unified “Motion + Conditional” representation.
To further enhance the locally aligned features in the decoder, a Cross-Transformer Refinement (CTR) module is introduced to adaptively aggregate cross-frame information, improving local consistency and preserving fine-grained geometric details.
In addition, a non-sequential hierarchical encoding paradigm is introduced to select both past and future reference frames, enabling more adaptive temporal context utilization and improving compression efficiency.

\subsection{Point Cloud Scene Flow Estimation}
Scene flow estimation extends 2D optical flow to 3D point clouds, enabling motion vector (MV) estimation in three-dimensional space. FlowNet3D~\cite{liu2019flownet3d} leverages PointNet++~\cite{qi2017pointnet++} for feature extraction and fusion, while PointPWC-Net~\cite{wu2020pointpwc} adopts a coarse-to-fine strategy with cost volumes. Pv-Raft~\cite{wei2021pv} introduces point-voxel correlation fields to capture both local and long-range point dependencies. Although effective, these methods are computationally expensive on large, dense point clouds due to their high dimensionality. In large-scale sequences, motion is subtle but significant in localized areas, where existing methods struggle with heterogeneous patterns due to non-uniform distributions. Our framework redefines motion modeling as a spatiotemporal alignment task by implicitly learning motion correlations in the feature space through lightweight point-based transformations. We introduce a latent space feature prediction module, anchoring current frame coordinates and using KNN~\cite{wu2019pointconv} to correlate reference features, predicting current frame features without 3D flow vectors, enhancing adaptability to localized deformations.

\subsection{Learned Video Compression}
% 最近，Learned video compression~\cite{hu2021fvc, hu2022coarse, lu2019dvc} 取得了优异的压缩性能，在2D视频的压缩上，Context-based frameworks~\cite{li2021deep, li2023neural, li2024neural} 的性能要明显优于“Motion + Residual” 框架~\cite{hu2021fvc, hu2022coarse, lu2019dvc}.  
% 最近，DCVC-RT~\cite的框架更是能让video实现实时的压缩，其采用里间接估计视频里的运动位移，大大减少了motion估计和motion补偿的
% 的计算复杂度。对于动态点云的压缩框架，和视频压缩框架存在相似性，也存在明显差异。由于点云的irregular geometry 
% 和复杂的非刚性运动，使得点云的运动估计更加具有挑战性，同视频压缩相似，采用“motion+Conditional”的点云压缩压缩框架取得了
% 非常险显著的进步，对比“Motion + Residual” 的压缩框架。针对这种差异性和相似性，我们采用了B—FMT方法，鲁棒的建模动态点云的
% 运动，并将预测的特征作为长下文，结合“motion+Conditional”框架，实现了我们自己的高效的动态点云压缩方法

% 最近，Learned video compression~\cite{hu2021fvc, hu2022coarse, lu2019dvc} 取得了优异的压缩性能，在2D视频的压缩上，Context-based frameworks~\cite{li2021deep, li2023neural, li2024neural} 的性能要明显优于“Motion + Residual” 框架~\cite{hu2021fvc, hu2022coarse, lu2019dvc}.  
% 最近，DCVC-RT~\cite{jia2025towards}的框架更是能让video实现实时的压缩，DCVC-RT采用间接估计视频里的运动位移的方法，大大减少了motion估计和motion补偿的
% 的计算复杂度。对于动态点云的压缩框架，和视频压缩框架存在相似性，也存在明显差异。由于点云的irregular geometry 
% 和复杂的非刚性运动，使得点云的运动估计更加具有挑战性，同视频压缩相似，采用“motion+Conditional”的点云压缩压缩框架~\cite{zhang2025adadpcc,2025A}取得了
% 非常险显著的进步，对比“Motion + Residual” 的压缩框架~\cite{fan2022d,akhtar2024inter}。针对这种差异性和相似性，我们采用了B—FMT方法，隐式的鲁棒并且快速建模动态点云的
% 运动,algin双向参考特征得到预测特征，并将预测的特征作为长下文，结合“motion+Conditional”框架，实现高效的动态点云压缩方法

Recently, learned video compression~\cite{hu2021fvc, hu2022coarse, lu2019dvc} has achieved remarkable progress. In the domain of 2D video compression, ``motion + conditional'' frameworks~\cite{li2024neural} significantly outperform traditional ``motion + residual'' schemes~\cite{hu2021fvc, hu2022coarse, lu2019dvc}. In particular, the DCVC-RT framework~\cite{jia2025towards} enables real-time video compression by implicitly estimating motion displacements, thereby greatly reducing the computational complexity of motion estimation and compensation.

Dynamic point cloud compression follows a similar paradigm, where motion cues are leveraged to guide conditional prediction. However, unlike videos defined on regular grids, point clouds exhibit irregular geometry and complex non-rigid motion, which make motion estimation considerably more challenging. Despite these difficulties, recent ``motion + conditional'' point cloud frameworks~\cite{zhang2025adadpcc,2025A} have demonstrated significant improvements over conventional ``motion + residual'' schemes~\cite{fan2022d,akhtar2024inter}, mirroring the evolution observed in video compression.

Motivated by these observations, we adopt a Bi-FMT strategy to implicitly model point cloud motion in a robust and efficient manner. The aligned features produced by Bi-FMT are first utilized as long-term context within a “motion + conditional” framework to guide predictive coding. Subsequently, at the decoding stage, we further refine these locally aligned representations through a Cross-Transformer Refinement (CTR) module, which adaptively aggregates cross-frame information to enhance local consistency and preserve fine-grained spatial details. This design ultimately enables efficient dynamic point cloud compression.
% The Contextual Decoder takes $X^4_t$ and the motion-aware context as input and outputs the decoded feature $\hat{F}^{3_\text{aglined}}_t$.
% This feature is further refined by the CBR module, yielding the refined representation $\hat{F}^{3_{\text{refined}}}_t$, which is then combined with the losslessly decoded coordinates $C^3_t$ to form the point cloud $\hat{X}^3_t$.

\begin{figure*}[h]
\centering
\includegraphics[width=1\textwidth]{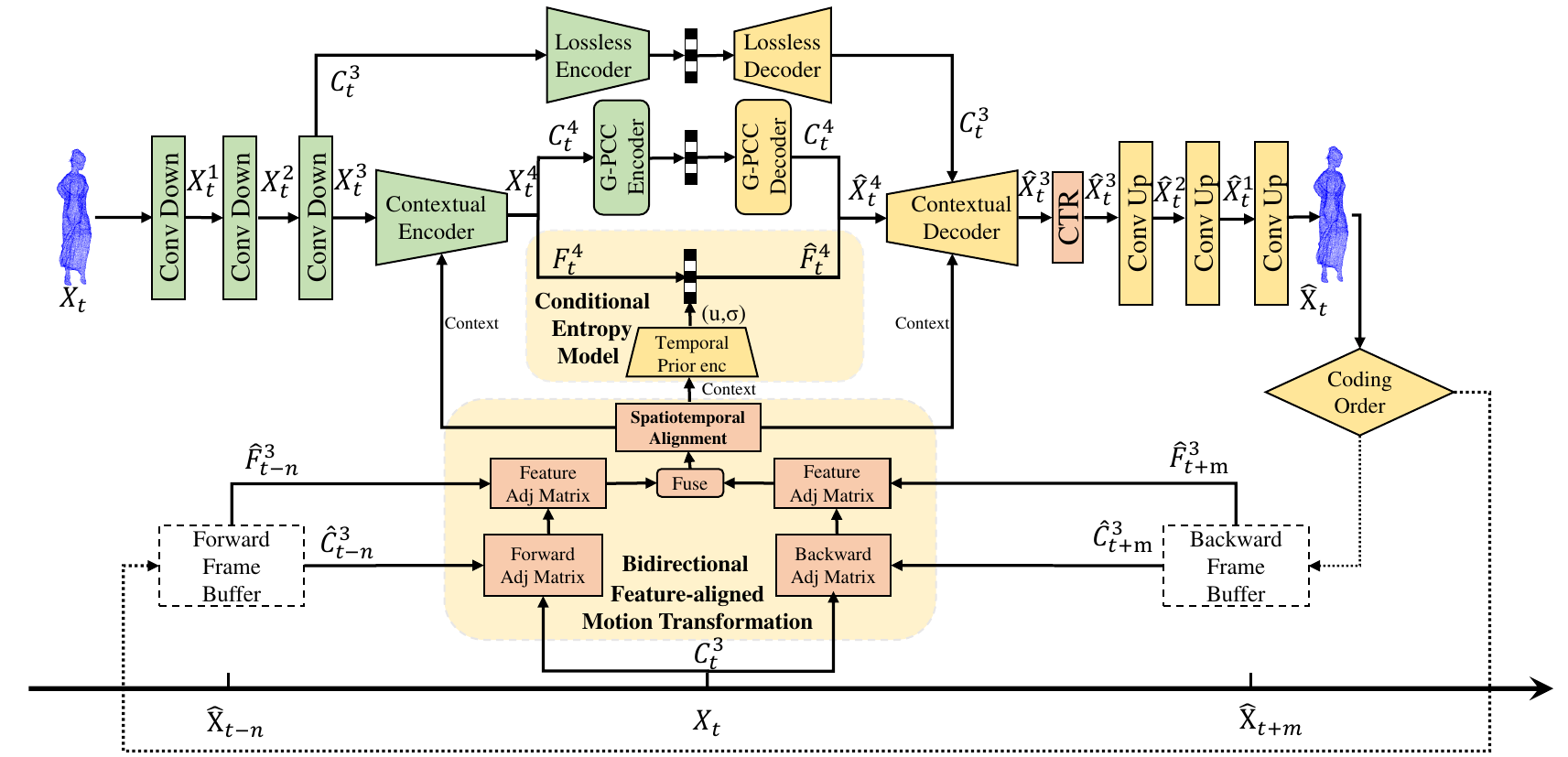} % Reduce the figure size so that it is slightly narrower than the column. Don't use precise values for figure width.This setup will avoid overfull boxes.

\caption{The framework employs a non-sequential hierarchical encoding mode. The input point cloud \( X_t \) is downsampled via three sparse convolutions to produce \( X^3_t \), consisting of coordinates \( C^3_t \) and features \( F^3_t \). These features are processed with reference point clouds $\hat{X}_{t-n}$ at time $t-n$ and $\hat{X}_{t+m}$ at time $t+m$ through a Bi-FMT module to generate temporally aligned features, which serve as temporal context for latent feature modeling. The downsampled coordinates $C^3_t$ are encoded using a learnable lossless codec, the coordinates $C^4_t$ extracted from the output $X^4_t$ of the Contextual Encoder are losslessly encoded using G-PCC, while the coordinate-related feature $F^4_t$ is encoded in a lossy manner using the Conditional
Entropy Model and subsequently reconstructed as $\hat{F}^4_t$. The Contextual Decoder takes $X^4_t$ and the motion-aware context as input and outputs the decoded feature $\hat{F}^{3_\text{aglined}}_t$.
This feature is further refined by the CTR module, yielding the refined representation $\hat{F}^{3_{\text{refined}}}_t$, which is then combined with the losslessly decoded coordinates $C^3_t$ to form the point cloud $\hat{X}^3_t$. Through three consecutive upsampling steps, $\hat{X}^3_t$ is reconstructed to the same scale as the original point cloud, resulting in $\hat{X}_t$. The decoded point cloud is then stored in either the forward or backward frame buffer.}
\label{fig:framework}
\end{figure*}

\section{Methodology}
This section first presents an overview of the proposed Bi-FMT framework. The subsequent subsections detail the encoder–decoder pipeline, feature extraction and reconstruction modules. The core components, including the Bidirectional Feature-aligned Motion Transformation and the Cross-Transformer for Fine-Grained Feature Alignment, are then elaborated to explain the implicit motion modeling and feature refinement processes. Finally, the conditional entropy model, the random access (RA) reference strategy, and the loss function are described to complete the overall framework.

 % \vspace{-10pt}

\subsection{Overview}
We represent the input dynamic point clouds as a sequence $X = \{X_0, X_1, \dots, X_t\}$, 
where each frame $X_t = (C_t, F_t)$ at time $t$ consists of 3D coordinates 
$C_t \in \mathbb{R}^{N \times 3}$ and associated features 
$F_t \in \mathbb{R}^{N \times d}$, representing voxel occupancy.
The overall compression framework is shown in Figure~\ref{fig:framework}.
Each frame is progressively downsampled into multi-scale latent representations, after which the lowest-resolution coordinates and reference frames from both past and future are fed into the Bi-FMT module to generate motion-aware context. 
The fused features are encoded by the Contextual Encoder and compressed using a conditional entropy model. 
At the decoder, Bi-FMT reconstructs the aligned context, which is first processed by the Contextual Decoder to produce aligned features, then refined by a bidirectional CTR aggregating forward and backward reference frames, and finally progressively upsampled to produce the output $\hat{X}_t$, stored according to the RA strategy for bidirectional reference.

% The overall compression framework is shown in Figure~\ref{fig:framework}. At the encoder side, the input point cloud is progressively downsampled, 第三次下采样的Coordinate采用lossless codec.接着经过Context Encoder,得到要编码传输的坐标和坐标对应的特征. 坐标采用G-PCC 编码器无损编码，特征采用Conditional Entropy Model进行熵编码。3组码流传输到解码端，在解码端，point clouds are reconstructed from the bitstream following the RA coding order, with decoded frames buffered as future references. 接下来我们详细描述下整个框架的编码过程和解码过程。

% and the resultant coordinates are losslessly compressed using a hybrid codec based on entropy coding and octree representation. The coordinates of the current frame and bidirectional references are then aligned via the Bi-FMT module to produce latent features, which serve as spatiotemporal context for the Conditional Entropy Model. At the decoder side, point clouds are reconstructed from the bitstream following the RA coding order, with decoded frames buffered as future references.

% \subsection{Encoder and Decoder}
\subsubsection{Encoding pipeline}
% \noindent\textbf{Encoding pipeline.}  
At time $t$, the input point cloud is denoted as 
$X_t = \{(C_t, F_t)\}$, 
where $C_t \in \mathbb{R}^{N\times 3}$ are the coordinates and 
$F_t \in \mathbb{R}^{N\times d}$ are the associated voxel occupancy features. 
The point cloud is progressively downsampled through three Downsampling Blocks, 
yielding $X^1_t$, $X^2_t$, and $X^3_t$, with 
$X^k_t = (C^k_t, F^k_t)$ denoting the stage-$k$ latent representation of frame $t$.  
The downsampled coordinates $C^3_t$, together with reference frames $\{\hat{X}_{t-n}, \hat{X}_{t+m}\}$ stored in the buffer, 
are fed into the Bidirectional Feature-aligned Motion Transformation (Bi-FMT) module, 
which aligns the reference features to the current frame and outputs the motion-aware context 
$F^{3_{\text{aligned}}}_t$. 
The concatenation of $F^3_t$ and $F^{3_{\text{aligned}}}_t$ is then encoded by the Contextual Encoder, 
producing a higher-level latent representation $X^4_t = (C^4_t, F^4_t)$.  

For compression, different strategies are adopted for different types of content:  
(i) $F^4_t$ is compressed by the Conditional Entropy Model;  
(ii) $C^4_t$ is losslessly encoded by the octree-based G-PCC geometry codec;  
(iii) $C^3_t$ is compressed by a learned end-to-end lossless geometry codec (similar to DDPCC~\cite{fan2022d}) 
to preserve the structural details required for alignment at the decoder side.  
\subsubsection{Decoding pipeline}
% \noindent\textbf{Decoding pipeline.}  
At the decoder side, $C^3_t$ is first recovered from the learned lossless codec, 
while $C^4_t$ is reconstructed by the G-PCC octree decoder. 
The latent features $\hat{F}^4_t$ are simultaneously obtained from the entropy decoder, 
leading to the high-level latent representation 
$\hat{X}^4_t = (C^4_t, \hat{F}^4_t)$.  

The Bi-FMT module then aligns the reference features using the decoded coordinates 
to reconstruct the context $X^{3_{\text{aligned}}}_t$. 
% Both $\hat{F}^4_t$ from $\hat{X}^4_t$ and $F^{3_{\text{aligned}}}_t$ from $X^{3_{\text{aligned}}}_t$ are incorporated by the Contextual Decoder, 
% which generates the temporally aligned features $\hat{F}^3_t$.  接着$\hat{F}^3_t$被双向的CTR模块进一步refine，得到refined feature $\hat{F}^{3_{refine}}_t$
% Combining $\hat{F}^3_t$ with $C^3_t$ forms the latent point cloud 
% $\hat{X}^3_t = (C^{3_{refine}}_t, \hat{F}^3_t)$. 
% Finally, $\hat{X}^3_t$ is progressively upsampled to obtain 
% $\hat{X}^2_t$, $\hat{X}^1_t$, and the full-resolution reconstruction $\hat{X}_t$, 
% which is stored in the reference buffer according to the RA coding order. 
Both $\hat{F}^4_t$ from $\hat{X}^4_t$ and $F^{3_{\text{aligned}}}_t$  from $X^{3_{\text{aligned}}}_t$ are integrated by the Contextual Decoder to generate the temporally aligned features $\hat{F}^{3_{\text{aligned}}}_t$.
The aligned features $\hat{F}^{3_{\text{aligned}}}_t$ are then further refined by a bidirectional CTR module, yielding the refined representation $\hat{F}^{3_{\text{refined}}}_t$.
Combining $\hat{F}^{3_{\text{refined}}}_t$ with $C^3_t$ produces the latent point cloud
$\hat{X}^3_t = (C^3_t, \hat{F}^{3_{\text{refined}}}_t)$.
Finally, $\hat{X}^3_t$ is progressively upsampled to obtain $\hat{X}^2_t$, $\hat{X}^1_t$, and the full-resolution reconstruction $\hat{X}_t$,
which is stored in the reference buffer according to the RA coding order.

\begin{figure}[t]
\centering
\includegraphics[width=1\columnwidth]{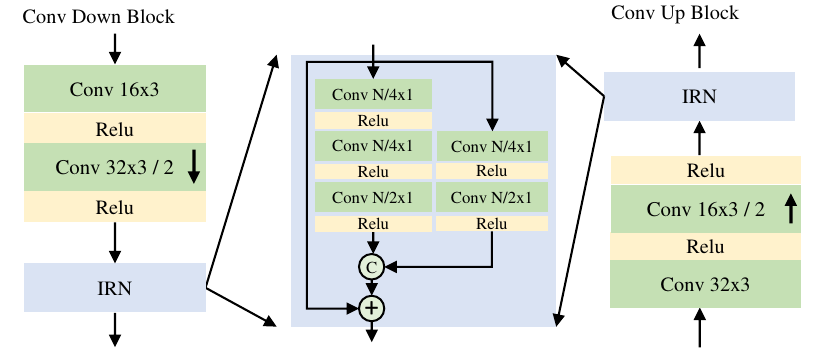} 
\caption{Architectural details of the upsampling and downsampling modules.}
\label{fig:upanddown}
\end{figure}

\subsection{Feature Extraction and Reconstruction Modules}

The basic modules of Bi-FMT-DPCC include downsampling/upsampling blocks, a contextual encoder, and a contextual decoder, forming the encoder–decoder backbone for feature extraction and reconstruction. Figure~\ref{fig:upanddown} illustrates the architecture of the point cloud downsampling and upsampling modules. Here, 'Conv $c{\times}n$' denotes a sparse convolution with $c$ output channels and a kernel size of $n{\times}n{\times}n$, implemented using the MinkowskiEngine (ME) library~\cite{choy20194d}. '$s{\uparrow}$' and '$s{\downarrow}$' represent upsampling and downsampling operations with a factor of $s$, respectively, while 'ReLU' denotes the Rectified Linear Unit and 'IRN' refers to the Inception-Residual Network~\cite{fan2022d} used for efficient feature aggregation.

Building upon these basic blocks, the Contextual Encoder (CE) integrates the downsampled input point cloud $X^3_t$ with bidirectionally aligned motion-aware context from Bi-FMT to produce enhanced features capturing both local and contextual cues. Similarly, the Contextual Decoder (CD) extends the upsampling block by fusing the reconstructed feature $\hat{X}^4_t$ with the same context, enabling more accurate feature reconstruction. Formally, these operations can be expressed as:
\begin{equation}
\begin{aligned}
X^4_t=f_\text{CE}\big(X^3_t,\, \text{Context}\big) &= \text{DownBlock}\big(X^3_t \oplus \text{Context}\big), \\
\hat{X}^3_t=f_\text{CD}\big(\hat{X}^4_t,\, \text{Context}\big) &= \text{UpBlock}\big(\hat{X}^4_t \oplus \text{Context}\big),
\end{aligned}
\end{equation}
where $\text{Context}$ denotes the bidirectionally aligned motion-aware features produced by Bi-FMT, and $\oplus$ represents concatenation.

\subsection{Bidirectional Feature-aligned Motion Transformation}

% In dynamic point cloud compression, modeling spatiotemporal correlations between reference and target frames is essential for effective inter-frame prediction. Due to the irregular and unordered nature of point clouds, explicit point-wise motion estimation is often unreliable and computationally unstable. Previous methods~\cite{fan2022d,xia2023learning,zhang2025adadpcc} typically rely on iterative motion refinement, where the KNN adjacency matrix is updated at each step. However, this continual change in neighborhood structure introduces significant variability, making the motion training process unstable and increasing the difficulty of learning accurate motion fields. This issue is further exacerbated under non-rigid motion, where irregular spatial deformations and numerous subtle displacements make gradient propagation highly sensitive, thereby hindering effective convergence. To mitigate these challenges, we reformulate the spatiotemporal motion compensation task as a coordinate-associated feature alignment problem, where the features are derived from voxel occupancy and are inherently correlated with the 3D coordinates.

In dynamic point cloud compression, explicit point-wise motion estimation is often unreliable due to the irregular and unordered structure of point clouds. Existing methods~\cite{fan2022d,xia2023learning,zhang2025adadpcc} iteratively refine motion using dynamically updated KNN graphs, which introduce instability and hinder convergence, especially under non-rigid motion with irregular deformations. To address this, we reformulate motion compensation as a coordinate-associated feature alignment task, where voxel-derived features are inherently correlated with 3D coordinates.

\begin{figure}[h]
\centering
\includegraphics[width=1\columnwidth]{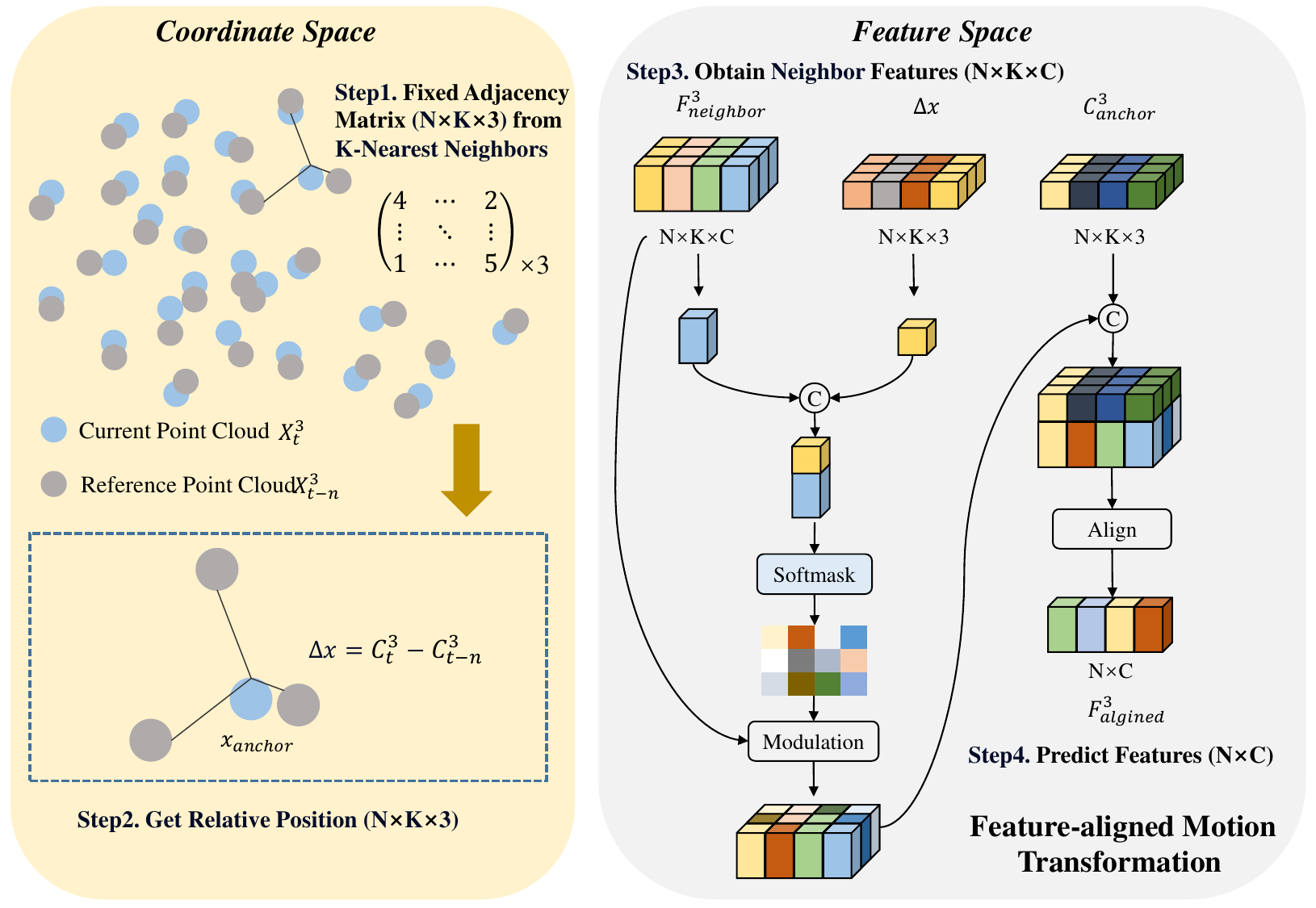} 
\caption{Detailed Illustration of the Unidirectional Feature-aligned
Motion Transformation for Spatiotemporal Alignment.}
\label{fig:FMT}
\end{figure}

We describe the feature alignment process using single-direction FMT, which references forward features once, as illustrated in Figure~\ref{fig:FMT}. First, in the coordinate space of the current frame $X^3_t$, a fixed KNN adjacency matrix is constructed between $X^3_t$ and the reference frame $X^3_{t-n}$. Based on this adjacency matrix, relative positions and neighbor features are extracted and aggregated. The aggregated features are then passed through a softmax function to produce a learnable, dynamically varying soft mask with values in [0,1]. This soft mask modulates the neighbor features $F^3_\text{neighbor}$ from $X^3_{t-n}$, resulting in weighted features, which are concatenated with the current coordinates $C^3_\text{anchor}$ along the channel dimension.

\begin{figure}[h]
\centering
\includegraphics[width=0.8\columnwidth]{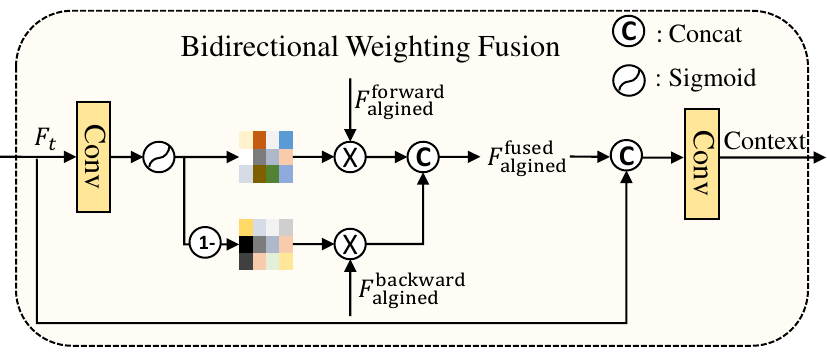} 
\caption{Illustration of the bi-directional weighting-based feature fusion.}
\label{fig:fused}
\end{figure}
Finally, a lightweight MLP rearranges the modulated features to obtain the aligned feature representation for the current frame. Notably, the fixed KNN adjacency matrix ensures a more stable training process compared to approaches that update the adjacency at each iteration. In our framework, we extend this approach using a Bidirectional Feature-aligned Motion Transformation (Bi-FMT), which incorporates both past and future reference frames to capture more complete temporal dependencies and subtle local deformations than single-direction FMT, thereby improving motion modeling for dynamic point clouds. The bidirectional feature alignment is conducted via two passes of Algorithm~\ref{alg:bi_fmt}, where the source reference frames $X_{ref}$ are set to $X_{t-n}$ and $X_{t+m}$ for backward and forward propagation, respectively. 
The fusion mechanism of forward-aligned features and backward-aligned features is illustrated in Figure~\ref{fig:fused}. We employ a weighted fusion approach, with the input and output of the fusion process defined by the following equations:
\begin{equation}
\begin{gathered}
F^{fused}_{aligned} = F_t \concat \left( w^{f}_t \odot F^{forward}_{aligned} \right) \odot \left( w^{b}_t \odot F^{backward}_{aligned} \right), \\
\text{with}\quad w^{f}_t = \sigma(Conv(F_t)), \\
\text{with}\quad w^{b}_t = 1 - w^{f}_t, \\
Context = Conv(F^{fused}_{algined}),
\end{gathered}
\end{equation}
\noindent
where $\text{Conv}$ denotes a convolutional layer, $\concat$ represents channel-wise concatenation, and $\odot$ indicates element-wise multiplication. $\sigma$ is the sigmoid activation function. $w^{f}_t$ and $w^{b}_t$ represent the forward and backward fusion weights at time $t$, respectively. Through end-to-end optimization, the learned fusion weights adaptively adjust the contributions of bidirectional temporal contexts based on their relevance to the target point cloud, enabling selective temporal fusion that effectively suppresses redundant or noisy information.
\begin{algorithm}[h]
\caption{: Bidirectional Feature Feature-aligned Motion Transformation}
\label{alg:bi_fmt}
\textbf{Input}: point cloud \(X_t = \{x_i^t\}_{i=1}^{N_{origin}}\), forward reference point cloud \(X_{t-n} = \{x_j^{t-n}\}_{j=1}^{M_{origin}}\) and backward reference point cloud \(X_{t+m} = \{x_j^{t+m}\}_{j=1}^{M_{origin}}\) \\
\textbf{Forward FMT}: 

\begin{algorithmic}[1]
\STATE Downsample \(X_t\) and \(X_{t-n}\) to obtain coordinate sets \(C^3_t = \{c_i^t\}_{i=1}^N\), \(C^3_{t-n} = \{c_j^{t-n}\}_{j=1}^M\) and associated feature \(F^3_t = \{f_i^t\}_{i=1}^N\), \(F^3_{t-n} = \{f_j^{t-n}\}_{j=1}^M\).
\FOR{each $c_i^t \in C_t$} \label{line:point_loop}
\STATE Find \(K\) nearest neighbors in \(C^3_{t-n}\):
$
\mathcal{N}_i = \{ c_{j_k}^{t-n} \mid j_k = \arg\min_{j} \| c_i^t - c_j^{t-n} \|, k=1,\dots,K \}
$
, then get a fixed adjacency matrix \( \mathrm{Adj} \in \mathrm{R}^{N \times K \times 3} \).
\STATE Extract neighbor features \(F^3_{\mathcal{N}_i} = \{f_{j_k}^{t-n}\}_{k=1}^K\) and compute relative coordinates:
$
\Delta x_i = \{ c_i^t - c_{j_k}^{t-n} \}_{k=1}^K
$.
\STATE Concatenate feature and relative coordinates 
$
H_i = \mathrm{concat}(F^3_{\mathcal{N}_i}, \Delta x_i) \in \mathrm{R}^{K \times (C + 3)}
$, then compute softmask via 
$
\alpha_i = \mathrm{Softmax}(MLP(H_i)) \in \mathrm{R}^{K \times 1}
$ and modulate neighbor features by 
$
\widetilde{F}^3_{\mathcal{N}_i} = \alpha_i \odot F^3_{\mathcal{N}_i}
$.
\STATE Concatenate modulated features and anchor coordinate:
$
Z_i = \mathrm{concat}\big( \widetilde{F}^3_{\mathcal{N}_i}, x_i \big) \in \mathrm{R}^{K \times (C + 3)}
$.
\STATE Apply MLP Networks on \(Z_i\) to predict feature for point \(x_i\):
$
F^3_{aligned} = \mathrm{MLP}(Z_i) \in \mathrm{R}^{C}
$, and then collect all predicted features:
$
F^{forward}_{3_{aligned}} = \{ F^3_{aligned} \}_{i=1}^N \in \mathrm{R}^{N \times C}
$.
\ENDFOR
\end{algorithmic}
\textbf{Backard FMT}:\({F}^{backward}_{3_{aligned}}\) =  \text{FMT}($X_t$, $X_{t+m}$) \\
\textbf{Output}: predicted features \({F}^{forward}_{3_{aligned}}\) and \({F}^{backward}_{3_{aligned}}\) for \(X_t\)
\end{algorithm}

Our proposed framework eliminates the need for transmitting motion-related bitstreams. 
Instead, the motion of dynamic point clouds is modeled through a Bidirectional Feature-aligned Motion Transformation (Bi-FMT) module at both the encoding and decoding stages. 
Specifically, KNN-based correspondence between $X_t$ and both past and future frames ($X_{t-n}$ and $X_{t+m}$) is first established to extract neighbor features and relative offsets, which are then interpolated and dynamically aligned via a lightweight MLP to produce motion-aware temporal context. 
By leveraging bidirectional references, Bi-FMT captures more complete temporal dependencies and subtle local deformations, leading to more accurate motion modeling compared to single-directional FMT. 
This motion-aware temporal context is then used for encoding, entropy modeling, and reconstruction.

Compared with other implicit motion methods like AuxGR~\cite{liu2024encoding}, which relies on KNN and self-attention (originally designed to address the limited receptive field of target convolution~\cite{akhtar2024inter}) and uses a "Motion + Residual" paradigm, the proposed Bi-FMT operates on a fixed KNN graph and learns a dynamically adjustable weight matrix to align features from forward and backward references. This modulation generates motion-aware context, making Bi-FMT better suited for non-rigid motions, while its "Motion + Conditional" representation informs the entropy model with aligned temporal context for more compact latent representations. By keeping the KNN graph fixed, Bi-FMT also enhances structural consistency and training stability.

\vspace{-7pt}
\subsection{Cross-Transformer for Fine-Grained Feature Alignment}
During the feature alignment stage in decoder, we employ Bi-FMT to achieve global feature alignment between reference and current representations. 
Although Bi-FMT effectively reconstructs global features, we observe that local regions still suffer from noticeable alignment artifacts, manifested as geometric distortions and local detail misalignments.
To further alleviate these local misalignment issues, we introduce a bidirectional, point-based Cross-Transformer Refinement (CTR) module in the decoder for fine-grained feature refinement. This module is designed to enhance the geometric consistency and local detail alignment of point features.

% Recent advances in Transformers and self-attention networks~\cite{vaswani2017attention} have achieved remarkable success in natural language processing and 2D image analysis. 
% Depending on how attention weights are modeled, self-attention mechanisms can be categorized into two types: 
% scalar attention~\cite{vaswani2017attention} and vector attention~\cite{zhao2020exploring,zhao2021point}. 
% Scalar attention is well-suited for global semantic modeling, 
% whereas vector attention is more appropriate for 3D point cloud tasks, 
% as it preserves both \emph{directionality} and \emph{channel dependency} among features, enabling the network to capture \emph{local geometric variations} effectively.
Inspired by Self-attention-based transformer~\cite{vaswani2017attention}, attention mechanisms are classified as scalar~\cite{vaswani2017attention} or vector~\cite{zhao2020exploring,zhao2021point} types. Vector attention better suits 3D point clouds by capturing local geometric variations via directional and channel-wise feature dependencies.
To enable fine-grained correspondence between two point clouds, 
we extend the vector attention formulation from self-attention~\cite{zhao2021point} 
to a cross-attention mechanism.
In our CTR module, the \emph{query} features are extracted from the aligned point cloud at time $t$, denoted as $X^{3_\text{aligned}}_t = (\hat{C}^3_t, \hat{F}^{3_\text{aligned}}_t)$,
while the \emph{key} and \emph{value} features are derived from the reference point cloud $X^{3_\text{ref}}_{t-n}$ and $X^{3_\text{ref}}_{t+m}$ .

\begin{figure}[t]
\centering
\includegraphics[width=1\columnwidth]{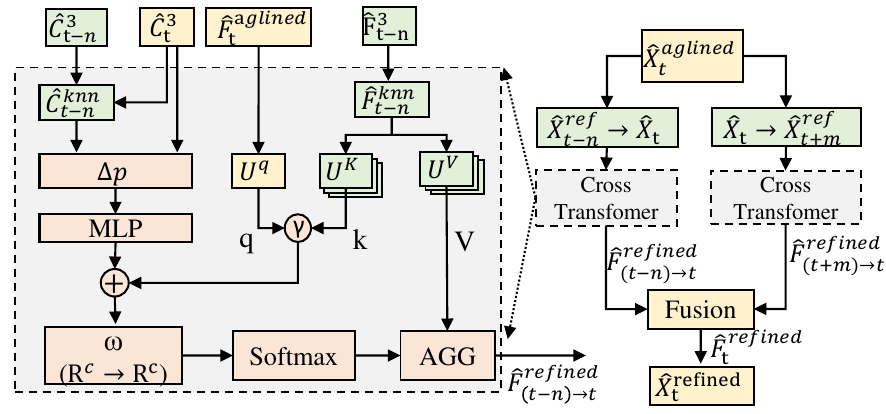} 
\caption{Detailed Illustration of the Cross Transformer for Local Feature Refinement, $\text{AGG}$ denotes the $\odot$ operation, which represents the channel-wise multiplication between the attention vector and the value feature, enabling feature-wise modulation. $\omega : {R}^c \mapsto {R}^c$ is a learnable weight encoder (e.g., an MLP) that computes the attention vectors to re-weight 
$\mathbf{v}_j$ across feature channels before aggregation. }
\label{fig:CTransfomer}
\end{figure}

Let $\mathbf{q}_i = \varphi(\mathbf{F}^{3_\text{aligned}}_i)$ denote the transformed feature of the $i$-th query point 
with spatial coordinate $\mathbf{C}^3_i$, 
and $\mathbf{k}_j = \psi(\mathbf{F}^{3_\text{ref}}_j)$, 
$\mathbf{v}_j = \alpha(\mathbf{F}^{3_\text{ref}}_j)$ 
denote the corresponding key and value features of the $j$-th reference point 
with coordinate $\mathbf{C}^{3_\text{ref}}_j$. $\phi$, $\psi$, and $\alpha$ are pointwise feature transformations, such as linear projections or MLPs.
For each query point $i$, we find its local neighborhood 
$\mathcal{N}(i)$ in the reference point cloud (e.g., $k$ nearest neighbors) 
and compute the cross-attention as:

\begin{equation}
\label{eq:cross-transformer}
\mathbf{y}^{3_\text{refined}}_i = 
\sum_{\mathbf{C}^{3_\text{ref}}_j \in \mathcal{N}(i)} 
\rho \Big(
\gamma \big( 
\mathbf{q}_i - \mathbf{k}_j + \delta(\mathbf{C}^{3_\text{aligned}}_i - \mathbf{C}^{3_\text{ref}}_j)
\big)
\Big) 
\odot 
\mathbf{v}_j
\end{equation}

where $\delta(\mathbf{C}^{3_\text{aligned}}_i - \mathbf{C}^{3_\text{ref}}_j)$ is a learnable positional encoding function
that captures the geometric relationship between query and reference points. 
The function $\gamma(\cdot)$ is implemented as a multilayer perceptron (MLP)
that outputs a vector attention weight for each feature channel, 
and $\rho(\cdot)$ denotes the softmax normalization across the $k$ neighbors. 
Finally, the aggregated feature $\mathbf{y}^{3_\text{cur}}_i$ is combined with the original feature $\mathbf{f}^{3_\text{cur}}_i$ 
via residual connection and layer normalization. $\odot$ denotes the channel-wise multiplication between the attention vector 
and the value feature, enabling feature-wise modulation. The bidirectional, point-based CTR layer is illustrated in Figure~\ref{fig:CTransfomer}. 
At time $t$, the decoded point cloud $\hat{X}^{3_\text{aligned}}_t$ takes the forward $\hat{X}^{3_\text{ref}}_{t-n}$ and backward $\hat{X}^{3_\text{ref}}_{t+m}$ point clouds at the same scale as references to perform Cross-Transformer operations. 
The refined features $\hat{F}^{3_\text{refined}}_{t-n}$ and $\hat{F}^{3_\text{refined}}_{t+m}$ are then fused in a weighted manner, following the same strategy as in Bi-FMT, 
to produce the fused features $\hat{F}^{3_\text{refined}}_t$. 
which is subsequently upsampled until the reconstructed point cloud reaches the same resolution as the original one.

This formulation allows each query point in the current frame 
to selectively aggregate context from its spatially corresponding region in the reference frame, 
enabling robust bidirectional feature refinement across frames.

\subsection{Conditional Entropy Model}

\begin{figure}[t]
\centering
\includegraphics[width=1\columnwidth]{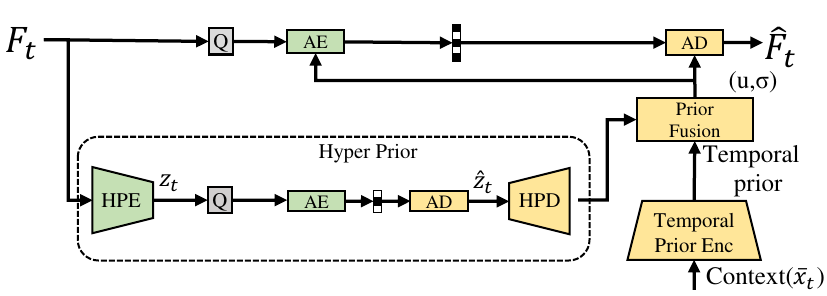} 
\caption{Our entropy model used to encode the quantized latent codes $\hat{F}_t$. HPE and HPD are hyper
prior encoder and decoder. Q means quantization. AE and AD are arithmetic encoder and decoder.}
\label{fig:cond_prior}
\end{figure}

We use the factorized entropy model for hyper prior and Laplace distribution to model the latent representations as\cite{li2021deep}. According to~\cite{shannon1948mathematical}, the cross-entropy between the estimated probability distribution and the actual latent code distribution is a tight lower bound of the actual bitrate, namely
\begin{equation}
R(\hat{y}_t) = \mathbb{E}_{\hat{y}_t \sim q_{\hat{y}_t}} \left[ \log_2 p_{\hat{y}_t}(\hat{y}_t) \right],
\end{equation}
where $p_{\hat{y}_t}(\hat{y}_t)$ and $q_{\hat{y}_t}(\hat{y}_t)$ denote the estimated and the true probability mass functions of the quantized latent codes $\hat{y}_t$, respectively. The gap between the actual bitrate $R(\hat{y}_t)$ and the bitrate estimated by cross-entropy is negligible. Therefore, our objective is to design an entropy model that can accurately estimate the probability distribution $p_{\hat{y}_t}(\hat{y}_t)$ of the latent codes. The framework of our entropy model is illustrated in Figure~\ref{fig:cond_prior} . First, we use the hyper prior model~\cite{balle2018variational} to learn the hierarchical prior and use auto regressive network~\cite{minnen2018joint} to learn the spatial prior. Meanwhile, we design a temporal prior encoder to explore the temporal correlation. The prior fusion network is trained to fuse three different priors and estimate the mean and scale of the latent code distribution. In this paper, we follow the existing work~\cite{kamisli2024dcc_vbrlic} and assume that $p_{\hat{F}_t}(\hat{F}_t)$ follows the Laplace distribution:
\begin{equation}
p_{\hat{F}_t}(\hat{F}_t \mid \overline{x}_t, \hat{z}_t) = \prod_i \mathcal{L}(\mu_i, \sigma_i) \cdot \mathcal{U}\left(-\frac{1}{2}, \frac{1}{2}\right)(\hat{F}_{t,i}),
\end{equation}
where
\begin{equation}
\mu_i, \sigma_i = f_{\text{pf}}\left(f_{\text{hpd}}(\hat{z}_t), f_{\text{ar}}(\hat{F}_{t,<i}), f_{\text{tpe}}(\overline{x}_t)\right).
\end{equation}
The index $i$ represents the spatial location. $f_{\text{hpd}}(\cdot)$ denotes the hyper prior decoder network, $f_{\text{ar}}(\cdot)$ is the autoregressive network, and $f_{\text{tpe}}(\cdot)$ refers to the specially designed temporal prior encoder network. $f_{\text{pf}}(\cdot)$ denotes the prior fusion network, the context $\overline{x}_t$ obtained from Bi-FMT serves as input to generate the temporal prior.

In our entropy model, $f_{\text{ar}}(\hat{F}_{t,<i})$ and $f_{\text{tpe}}(\overline{x}_t)$ provide the spatial and temporal priors, respectively, while $f_{\text{hpd}}(\hat{z}_t)$ provides the supplemental side information that cannot be learned from spatial or temporal correlations.

\subsection{Random Access (RA) Reference Strategy} 
\begin{figure}[t]
\centering
\includegraphics[width=0.9\columnwidth]{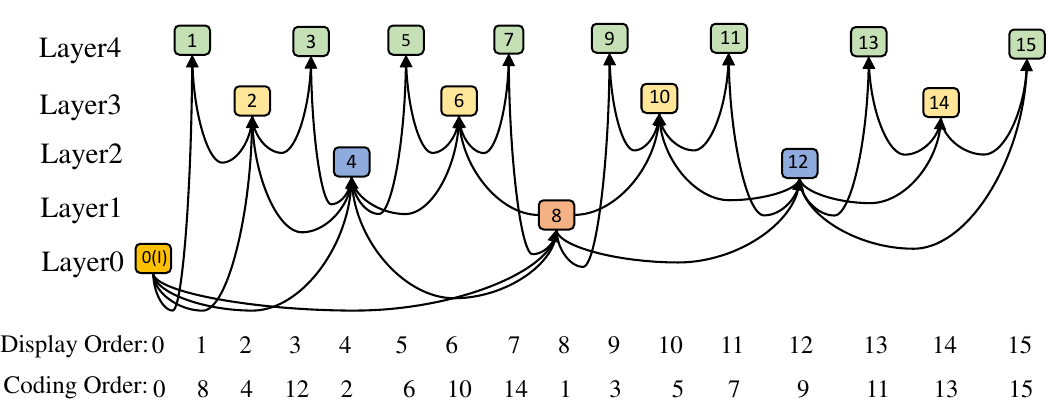}
\caption{Illustration of the hierarchical RA referencing structure for a GoF of 16. Frames are organized into five temporal layers and encoded in non-sequential order. Arrows indicate reference dependencies. Layer 0 (yellow) is the I-frame, while other frames serve as P/B references according to their layer. This strategy enables bidirectional prediction, random access, and efficient parallel encoding.}

\label{fig:RA}
\end{figure}

In Bi-FMT, the bidirectional temporal references are organized using a non-sequential, hierarchical Group of Frames (GOF) structure, which enhances temporal modeling and improves compression efficiency. Unlike conventional video codecs that periodically insert I-frames to mitigate drift, dynamic point cloud sequences exhibit stable coding quality across long durations. Thus, only the first frame in the entire sequence is intra-coded, while all subsequent frames are inter-coded.
Following the open-GOF configuration inspired by H.266~\cite{bross2021overview}, each GOF consists of 16 frames organized into five hierarchical layers and is encoded in a bottom-up, layer-wise manner. To maintain temporal continuity, the first frame of each GOF references the last frame of the preceding GOF, employing inter-frame coding. All frames are encoded in a non-sequential order to maximize temporal prediction efficiency, following the coding order in a GOF (i.e., 0, 8, 4, 12, 2, 6, 10, 14, 1, 3, 5, 7, 9, 11, 13, 15), as illustrated in Figure~\ref{fig:RA}. Since the reference frames for higher-layer frames are exclusively located in lower layers, frames within the same layer can be encoded and decoded in parallel, enabling efficient synchronization across the hierarchy~\cite{sullivan2012overview}.

Within each GOF, both unidirectional and bidirectional predictions are employed with precision based on spatiotemporal relevance. For instance, Frame 14 uses Frame 12 as its primary single reference due to its proximity, while Frame 10 utilizes bidirectional cues from Frames 8 and 12 for significantly enhanced prediction.

\subsection{Loss Function} 
Given an input point cloud $x$, the optimization objective of the network is shown in Equation~\ref{eq:rd}, where $\lambda$ is the Lagrange multiplier that controls the trade-off between bitrate and distortion. Here, $f(\cdot)$ and $g(\cdot)$ denote the encoder and decoder, respectively, and quantization is denoted by $\lfloor \cdot \rceil$:
\begin{equation}
\mathbb{E}_{x \sim p_x} \left[ -\log p_y\left( \left\lfloor f(x) \right\rceil \right) \right] + \lambda \cdot \mathbb{E}_{x \sim p_x} \left[ d\left(x, g\left( \left\lfloor f(x) \right\rceil \right) \right) \right]
\label{eq:rd}
\end{equation}
 
As described in Equation~\ref{eq:rd}, the loss function is modeled as a joint optimization of rate and distortion, i.e., $R + \lambda \cdot D$. Since $z_t$ serves as side information, its bitrate must also be included in the total rate calculation. Accordingly, the estimated rate is given by Equation~\ref{eq:rate}, while the distortion is computed using the Binary Cross Entropy (BCE) loss, following PCGCv2~\cite{bross2021overview}, as shown in Equation~\ref{eq:distortion}:

\begin{equation}
R = \mathbb{E}_{x \sim p_x} \left[ \sum_{t} \left( \log p(y_t \mid y_{t-1}, z_t) + \log p(z_t) \right) \right]
\label{eq:rate}
\end{equation}

\begin{equation}
D = \frac{1}{K} \sum_{k=1}^{K} \left( \frac{1}{N_k} \sum_{v} -\left(O_v \log p_v + (1 - O_v) \log(1 - p_v)\right) \right)
\label{eq:distortion}
\end{equation}

In Equation~\ref{eq:distortion}, $O_v$ denotes the ground-truth occupancy value for voxel $v$, $p_v$ is the predicted occupancy probability, and $N_k$ is the number of voxels considered at the $k$-th upsampling stage. To minimize distortion, BCE losses from all $K$ stages of the synthesis transform are averaged, yielding the final distortion metric.

\section{Experiments}
This section presents subjective and objective evaluations of the proposed method, followed by analyses of method efficiency and bitstream composition, and ablation studies on the contributions of individual modules.
\subsection{Implementation Details}
\subsubsection{Training Dataset}
Consistent with D-DPCC~\cite{fan2022d} and adaDPCC~\cite{zhang2025adadpcc}, the proposed model is trained on the Owlii dataset~\cite{owlii}, which comprises four sequences totaling 2,400 frames, with each sequence lasting 20 seconds at 30 frames per second (FPS). To optimize training efficiency, we quantize
the original 11-bit precision point cloud data to 10-bit.

\subsubsection{Evaluation Dataset}
Following the MPEG common test condition (CTC)~\cite{mpeg2022}, we evaluate the performance of the proposed D-DPCC framework using 8i Voxelized Full Bodies
(8iVFB)~\cite{d20178i}, containing 4 sequences with 1200 frames. The frame rate is 30 fps over 10 seconds.
We set the GoF size to 16. The first frame of each GoF is designated
as an “I” frame and encoded by SparsePCC~\cite{wang2022sparse} while subsequent frames are labeled
as “P” frames or "B" frames and encoded by the proposed method.
In the coding structure, ``P'' denotes uni-directional reference, and ``B'' denotes bi-directional reference.

\subsubsection{Experiment Setup}
We train Bi-FMT-DPCC with $\lambda$= \{1, 3, 5, 8,15\} for each rate point. We utilize an Adam~\cite{kinga2015method} optimizer with $\beta = (0.9, 0.999)$, together with a learning rate scheduler with a decay rate of $0.5$ for every $15$ epochs. A two-stage training strategy is employed for each rate point. In the initial stage, the model is trained for the first five epochs with $\lambda$ set to 30, which facilitates faster convergence of the point cloud reconstruction module. In the subsequent stage, training continues for an additional 45 epochs with $\lambda$ restored to its original value to fine-tune the model. Due to GPU memory limitations, a batch size of 1 is used, and all training is performed on an NVIDIA 4090 GPU. In our implementation, following the aforementioned settings, the number of neighbors $K$ in KNN is set to 32, and the GoF size is set to 16.
\subsubsection{Evaluation Metrics}
The bit rate is evaluated using bits per
point (bpp), and the distortion is evaluated using point-to-point geometry (D1) Peak Signal-to-Noise Ratio (PSNR) and
point-to-plane geometry (D2) PSNR following the MPEG
CTC. The peak value ${p}$ is set as 1023 for 8iVFB.

\subsection{Performance Evaluation}
The proposed Bi-FMT-DPCC is compared with the current state-of-the-art (SOTA) dynamic point cloud geometry compression framework, V-PCC Test Model v13, using quantization parameter (QP) values of 18, 15, 12, 10, and 8. Comparisons are also performed with SOTA deep learning-based frameworks for dynamic point cloud geometry compression, including D-DPCC~\cite{fan2022d}, patchDPCC~\cite{pan2024patchdpcc}, and AdaDPCC~\cite{zhang2025adadpcc}, which are trained on the Owlii dataset and evaluated on the 8iVFB dataset. In addition, PCGCv2~\cite{wang2021multiscale}, commonly employed for static point cloud compression, is included for reference.

\begin{table*}[ht]
    \centering
    \small
      \caption{BD-Rate gains measured using both D1 PSNR(\%) and D2 PSNR(\%) for IMNR-DPCC (ours) against the V-PCC, D-DPCC (Fan et al. 2022, IJCAI), PatchDPCC (Pan et al. 2024, AAAI), AdaDPCC (Zhang et al. 2025, AAAI) and PCGCV2 (Wang et al. 2021, DCC) for lossy coded point clouds. The larger the negative percentage, the greater the gain of our method.
    }
    \resizebox{\linewidth}{!}{
     \begin{tabular}{c c c c c c c c c c c} 
        \toprule
        {}&\multicolumn{2}{c}{\textbf{Gain over}}&\multicolumn{2}{c}{\textbf{Gain over}}&\multicolumn{2}{c}{\textbf{Gain over }}&\multicolumn{2}{c}{\textbf{Gain over}}&\multicolumn{2}{c}{\textbf{Gain over }}\\
        
        {}&\multicolumn{2}{c}{\textbf{VPCC}}&\multicolumn{2}{c}{\textbf{D-DPCC}}&\multicolumn{2}{c}{\textbf{PatchDPCC}}&\multicolumn{2}{c}{\textbf{AdaDPCC}}&\multicolumn{2}{c}{\textbf{PCGCV2}}\\
        \midrule
        \textbf{Dataset} & \textbf{D1 (\%)} & \textbf{D2 (\%)} & \textbf{D1 (\%)} & \textbf{D2 (\%)} &\textbf{D1 (\%)} & \textbf{D2 (\%)} & \textbf{D1 (\%)} & \textbf{D2 (\%)}& \textbf{D1 (\%)}& \textbf{D2 (\%)} \\ 
        \midrule
        % \multirow{2}{s} & 0 & 0 \\
        Longdress &  -68.24 & -70.81& -23.58 & -24.67& -19.18 & -46.96   &-13.95&-17.60& -43.43 & -60.92\\ 
        Loot      &  -70.07 & -90.63& -22.45 & -21.41& -17.18 & -32.07   &-9.55&-7.70& -49.08 & -60.03\\ 
        Redandblack &  -75.56 & -96.75& -15.44 & -20.22& -9.52 & -48.44   &-8.96&-20.91& -36.70 & -63.55\\ 
        Soldier     &  -86.12 & -98.26& -19.14 & -16.66& -8.87 & -38.36   &-5.15&-4.01& -52.78 & -58.81\\
    
        \hline
        Average& \textbf{-75.00} & \textbf{-89.11}& \textbf{-20.15} & \textbf{-20.74}& \textbf{-13.69} &  \textbf{-41.46}&\textbf{-9.40} &  \textbf{-12.55}& \textbf{-45.5} & \textbf{-60.83} \\
        \bottomrule
    \end{tabular}
    }
    % \caption{BD-Rate gains measured using both D1 PSNR(\%) and D2 PSNR(\%) for IMNR-DPCC (ours) against the V-PCC, D-DPCC~\cite{fan2022d}, PatchDPCC\cite{pan2024patchdpcc}, AdaDPCC~\cite{zhang2025adadpcc} and PCGCV2~\cite{wang2021multiscale} for lossy coded dense point clouds. The larger the negative percentage, the greater the gain of our method.
    % }
  
    \label{tab:rd_comparison}
\end{table*}

\begin{figure*}[t]
\centering
\includegraphics[width=1\textwidth]{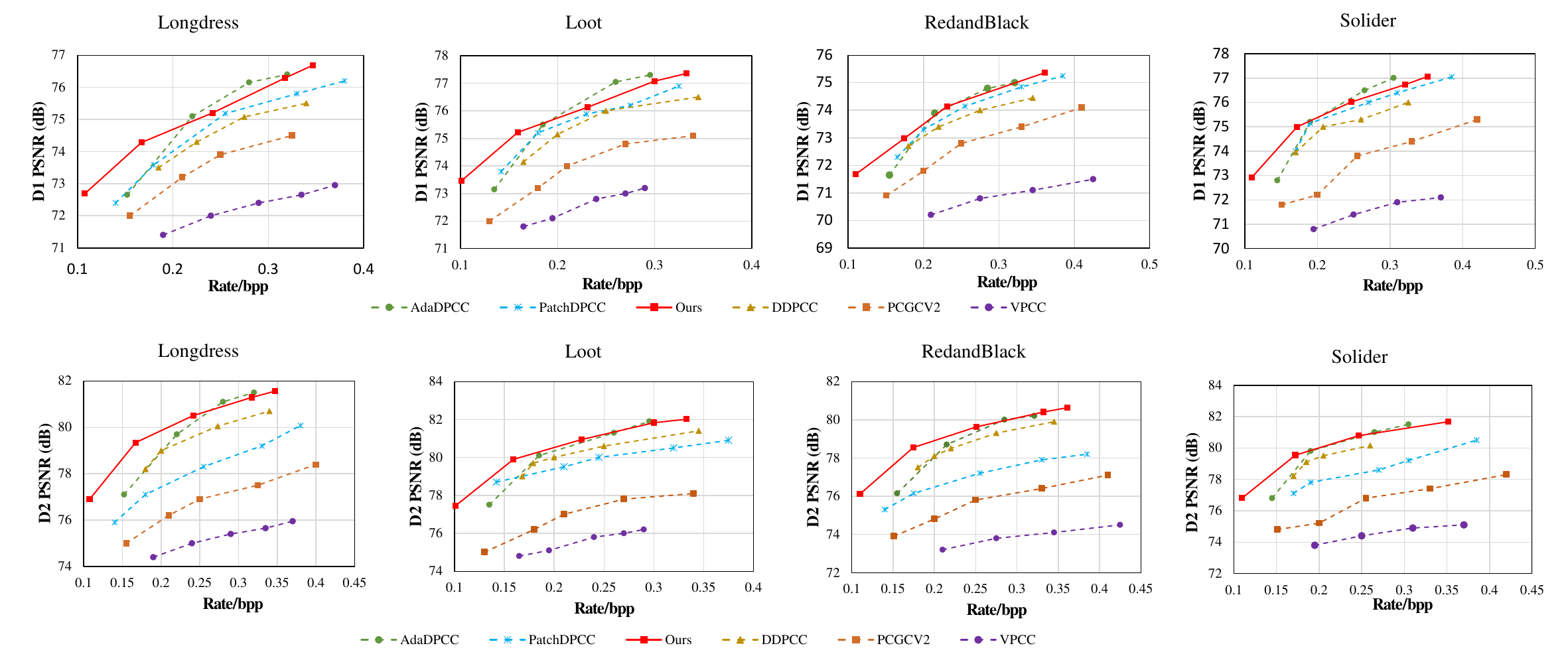}
\caption{D1-PSNR and D2-PSNR Rate-distortion curves of different compression methods on MPEG 8i. Please zoom in for more details.}
\label{fig:BD-PSNR}
\end{figure*}

\subsubsection{Quantitative comparison} The rate-distortion performance of different methods is shown in Figure~\ref{fig:BD-PSNR}, and the corresponding BD-Rate gains are summarized in Table~\ref{tab:rd_comparison}. Compared with the projection-based method like V-PCC, the proposed method targets 3D space compression with powerful learning-based modules and culminates in BD-Rate reduction exceeding 75.00\% (D1) and 89.11\% (D2) on
average. Furthermore, our method achieves an average BD-Rate reduction of over 45.5\% for D1 and 60.83\% for D2 relative to PCGCv2. Compared with other learning-based dynamic point cloud compression approaches~\cite{fan2022d,pan2024patchdpcc,zhang2025adadpcc}, our framework demonstrates superior performance, particularly in terms of the D2-PSNR metric. Notably, even when compared with the current state-of-the-art method AdaDPCC~\cite{zhang2025adadpcc}, our approach achieves significant BD-Rate reductions of 9.4\% for D1 and 12.5\% for D2.

It is worth noting that the performance gains over AdaDPCC become more prominent under low bitrate scenarios. This can be attributed to the global effectiveness of the bidirectional alignment process, which is particularly beneficial when bit resources are constrained, enabling more efficient representation of temporal dynamics.

\subsubsection{Qualitative evaluations}
\begin{figure}[ht]
\centering
\includegraphics[width=1\columnwidth]{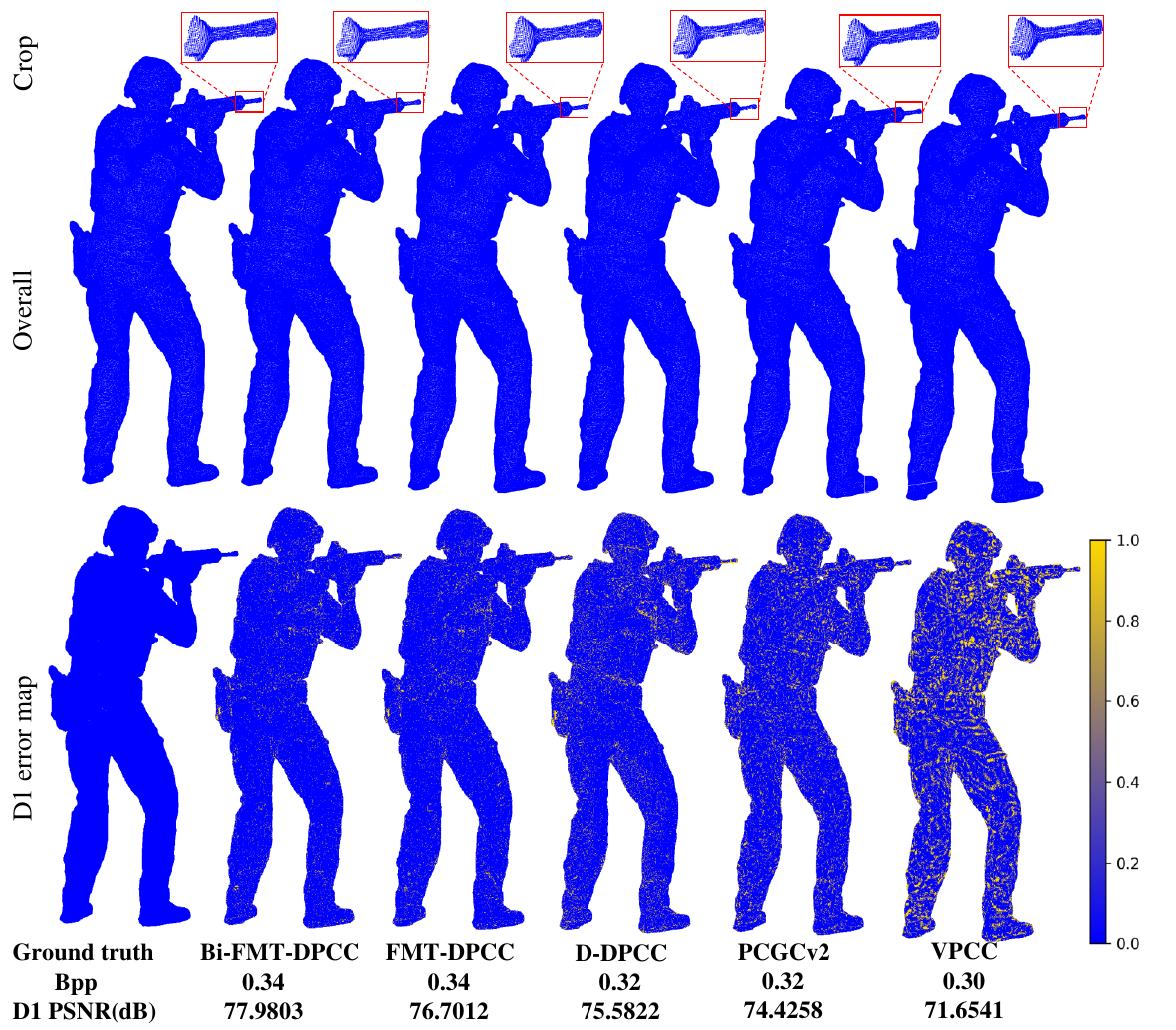} % Reduce the figure size so that it is slightly narrower than the column. Don't use precise values for figure width.This setup will avoid overfull boxes.
\caption{Visualization of geometry reconstruction results for the $Soldier$ sequence using different compression methods. 
Red rectangles in the first row indicate regions enlarged in the second row. 
The bottom row shows D1 error maps, where brighter (yellow) areas denote larger reconstruction errors.
}
\label{fig:visual_compare}
\end{figure}

We visualize the reconstructed point clouds obtained by different geometry compression methods. 
Figure~\ref{fig:visual_compare} presents the overall geometry of the entire point cloud, a zoomed-in region highlighting geometric details, and the corresponding error map in terms of D1 distance for the $Soldier$ sequence. 
At similar bitrates, the proposed Bi-FMT-DPCC exhibits significantly lower reconstruction errors than D-DPCC, particularly in the region where the soldier moves with the gun. 
This region involves complex non-rigid motion, which the motion estimation in 3DAWI-based D-DPCC fails to capture effectively, resulting in notable reconstruction errors. 
Compared to D-DPCC, the single-directional FMT-DPCC achieves more accurate reconstruction, benefiting from the implicit deformation modeling capability of the FMT, while the introduced Cross-Transformer Refinement (CTR) module further alleviates feature alignment errors. 
For the bidirectional variant, Bi-FMT, the error regions are further reduced, as it captures more complete temporal dependencies and subtle local deformations, leading to more precise motion modeling than its single-directional counterpart. 
In addition, when examining the enlarged local details of the gun, we observe that PCGCv2 and VPCC can reconstruct the object contours well but fail to recover fine-grained structures. 
In contrast, our method faithfully preserves high-fidelity geometric details, especially at the gun tip.

\subsection{Method Efficiency}

\subsubsection{Coding time comparison}
Table~\ref{tab:speed_table} presents a comprehensive comparison of average encoding and decoding times on the 8iVFB dataset, all evaluated under the same hardware and software environment using an Nvidia GeForce GTX 4090 GPU with CUDA 12.0. Our method demonstrates a significant advantage in both encoding and decoding efficiency over existing approaches. Specifically, our framework achieves an encoding time of 0.59s, which is 1.15$\times$ faster than AdaDPCC (0.68s) and 8.32$\times$ faster than PatchDPCC (4.91s). For decoding, our method requires only 0.57s, achieving 1.18$\times$ speedup over AdaDPCC (0.67s) and 2.09$\times$ speedup over PatchDPCC (1.19s). Note that V-PCC involves a projection process from 3D data to 2D and is executed primarily on CPUs, resulting in extremely long encoding times, therefore, it is not included in the comparison.

It is worth noting that our current implementation does not yet incorporate parallel acceleration based on the hierarchical reference structure. Since the hierarchical reference structure is inherently well-suited for parallel processing, we expect that both encoding and decoding speeds can be further improved with parallelization~\cite{sullivan2012overview}.

\begin{table*}[ht]
    \centering
    \footnotesize
     \caption{Average coding time comparison (encoding / decoding) on the 8iVFB dataset. Time is measured in seconds (s).}
    \setlength{\tabcolsep}{15pt}
    \renewcommand{\arraystretch}{1}{
    % \resizebox{\linewidth}{!}{
    % \begin{tabular}{p{1cm} p{1cm} p{1cm} p{1cm} p{1cm}} % 设置列宽
    \begin{tabular}{l c c c c c c c} % 设置列宽
        \toprule
        \textbf{Method} & {Soldier}& {Redandblack}& {Loot}& {Longdress}&\textbf{Average} \\ 
        \midrule
        Ours & \textbf{0.63 / 0.57} &\textbf{0.50 / 0.52} &\textbf{0.62 / 0.58} &\textbf{0.62 / 0.61} &\textbf{0.59 / 0.57}   \\
        % PCGCV2 & 1.07 / 0.83 &1.02 / 0.76&1.12 / 0.86 &1.14 / 0.89 &1.09 / 0.84   \\
        D-DPCC & 1.36 / 1.21 &1.29 / 1.13 &1.19 / 1.12 &1.26 / 1.16&1.28 / 1.16    \\
        PatchDPCC & 5.06 / 1.34 &4.75 / 1.12 &4.86 / 1.10 &4.98 / 1.20&4.912 / 1.19    \\
        AdaDPCC & 0.71 / 0.67  & 0.58 / 0.63 &0.69 / 0.68 &0.72 / 0.71&0.68 / 0.67  \\
        
        \bottomrule
    \end{tabular}
    }
    \label{tab:speed_table}
\end{table*}

\subsubsection{Module-wise Runtime}
Table~\ref{tab:module_time} summarizes the runtime distribution of different modules in the proposed Bi-FMT-DPCC framework. 
The overall processing time per frame is approximately 1085\,ms. 
Among all components, the Conditional Entropy Model dominates the runtime, accounting for 46\% of the total computation time. 
The Contextual Decoder and Upsampling modules together contribute 18.43\%, while the Bi-CTR and Bi-FMT modules in the decoder occupy 7.37\% and 11.06\%, respectively. 
The encoder part, including Downsampling and Contextual Encoder, takes around 5.99\% of the total runtime. 
This analysis indicates that the entropy model is the main computational bottleneck, whereas the proposed Bi-FMT and Bi-CTR modules add only minor overhead, reflecting the overall efficiency of the proposed framework.

\begin{table}[H]
    \centering
    \caption{Runtime analysis of individual modules in the proposed Bi-FMT-DPCC framework.}
    \resizebox{0.9\columnwidth}{!}{%
    \label{fig:module_time}
    \begin{tabular}{lcc}
    \toprule
    \textbf{Module} & \textbf{Runtime (ms)} & \textbf{Percentage (\%)} \\
    \midrule
    Downsampling + Contextual Encoder & 65   & 5.99 \\
    Conditional Entropy model  & 500  & 46.08 \\
    Bi-FMT Module (Encoder)            & 120   & 11.06 \\
    \midrule
    Contextual Decoder + Upsampling   & 200  & 18.43 \\
    Bi-CTR (Decoder)                            & 80  & 7.37\\
    Bi-FMT Module (Decoder)            & 120   & 11.06 \\
    \midrule
    \textbf{Total}                           & 1085 & 100.00 \\
    \bottomrule
    
    \end{tabular}
    }
    \label{tab:module_time}
\end{table}

% \vspace{-10pt}
\subsection{Bitstream Composition}
\begin{figure}[t]
\centering
\includegraphics[width=1\columnwidth]{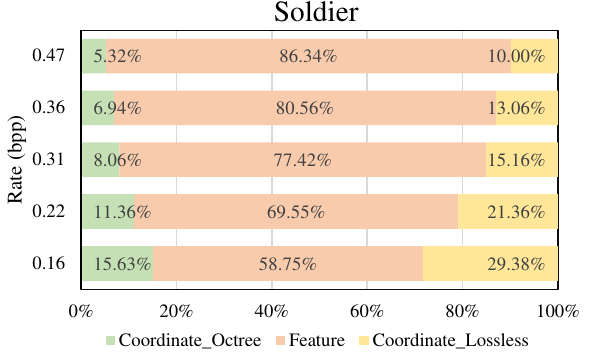} 
\caption{Bitstream composition of the \textit{Soldier} sequence at different bitrates, 
showing the proportions of \textit{Coordinate\_Lossless}, \textit{Coordinate\_Octree}, and \textit{Feature}.}
\label{fig:bitstream}
\end{figure}

To further investigate the bit allocation behavior and transmission efficiency of the proposed Bi-FMT-DPCC framework, 
we analyze the bitstream composition across different bitrate levels, as illustrated in Figure~\ref{fig:bitstream}. 
The transmitted bitstream primarily consists of three components: 
(1) \textit{Coordinate\_Lossless}, representing the downsampled coordinates $C^3_t$ encoded with a learnable lossless codec~\cite{fan2022d}; 
(2) \textit{Coordinate\_Octree}, corresponding to the octree-partitioned coordinates $C^4_t$ encoded by the G-PCC codec;
(3) \textit{Feature}, denoting the latent features $F^4_t$ compressed by the proposed conditional entropy model. 
As shown in Figure~\ref{fig:bitstream}, the proportion of feature bits (\textit{Feature}) consistently dominates the total bitrate across all compression levels, 
indicating that the majority of information resides in the latent feature domain. 
Meanwhile, the coordinate components (\textit{Coordinate\_Lossless} and \textit{Coordinate\_Octree}) occupy a relatively small and stable portion of the bitstream. 
At lower bitrates, the fraction of feature bits decreases slightly due to stronger entropy regularization, 
whereas at higher bitrates, more bits are allocated to feature representation to preserve fine geometric and texture details. 
This trend reflects a balanced trade-off between compactness and reconstruction fidelity, 
demonstrating that our framework effectively allocates bits according to the importance of feature and geometry components.

\begin{figure}[t]
\centering
\includegraphics[width=0.9\columnwidth]{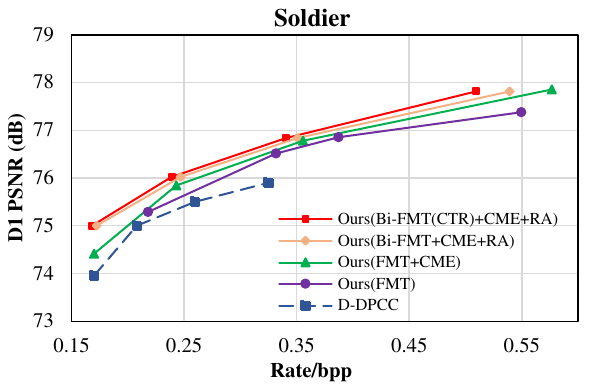} 
% \caption{Ablation Study on RD Performance: INMR denotes the Implicit Neural Motion Representation module, CME refers to the Conditional Entropy Model, and RA indicates the hierarchical coding structure under the random access configuration.}
\caption{Ablation study on RD performance. The contributions of major components are evaluated: Bi-FMT(CTR) models motion implicitly and refines local artifacts, CME (Conditional Entropy Model) improves compression with learned priors, and RA enables hierarchical non-sequential coding.}
\label{fig:ablation_study}
\end{figure}

\begin{table}[ht]
\centering
\caption{Ablation study on the effectiveness of proposed modules}
\label{tab:bdrate_ablation}
\begin{tabular}{lcccc}
\toprule
Method & $M_a$ & $M_b$ & $M_c$ & $M_d$  \\
\midrule
FMT       & $\checkmark$ & $\checkmark$  & $\times$ & $\times$ \\
Bi-FMT    & $\times$ & $\times$  & $\checkmark$& $\checkmark$ \\
Bi-CTR    & $\times$ & $\times$  & $\times$& $\checkmark$ \\
CME       & $\times$ & $\checkmark$  & $\checkmark$& $\checkmark$ \\
RA       & $\times$ & $\times$  & $\checkmark$ & $\checkmark$ \\
\midrule
BD-Rate(D1) & -9.86 & -12.27  & -16.55  & -19.14 \\
\bottomrule
\end{tabular}
\end{table}

\vspace{-10pt}
\subsection{Ablation Study}
\subsubsection{RD performance} 

To validate the effectiveness of our proposed modules, including the Bi-FMT module, the Conditional Entropy Model (CEM) based on Bi-FMT, the unordered hierarchical coding structure under the Random Access (RA) mode, and the CTR module at the decoder, we conducted extensive ablation studies.

\begin{figure}[h]
\centering
\includegraphics[width=0.8\columnwidth]{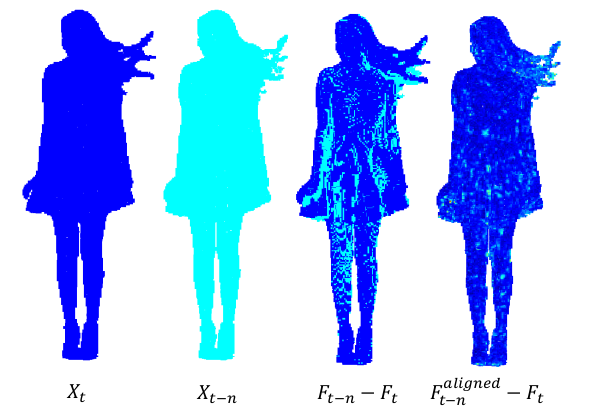} % Reduce the figure size so that it is slightly narrower than the column. Don't use precise values for figure width.This setup will avoid overfull boxes.
\caption{From left to right: point cloud at time \( t \), reference point cloud at time \( t{-}n \), feature difference before alignment, and reduced feature difference after alignment.}
\label{fig:aligned}
\end{figure}

Figure~\ref{fig:ablation_study} presents the RD curves of the \textit{Soldier} sequence. 
By examining these curves, it can be observed that progressively removing each module consistently degrades the rate-distortion performance, highlighting the importance of each component. 
To quantitatively compare the contributions of each module, Table~\ref{tab:bdrate_ablation} summarizes the BD-Rate performance of different model variants $M_{a–d}$. 
Model $M_a$, which includes only the FMT module, provides a significant improvement over the 3DAWI method in D-DPCC, achieving a BD-Rate reduction of approximately 9.86\%. 
Here, 3DAWI is an explicit KNN-based motion estimation method. 
By leveraging the FMT-aligned features as context and integrating them with the hyperprior to construct a hyperprior-context model, model $M_b$ achieves an additional BD-Rate gain of approximately 12.77\%. 
Model $M_c$ further incorporates Bi-FMT with the unordered hierarchical reference structure under the Random Access configuration, yielding an additional BD-Rate reduction of around 16.55\%. 
Finally, model $M_d$ builds upon this bidirectional feature alignment and integrates the CTR module at the decoder, which refines locally aligned features, enhances local consistency, and restores fine-grained spatial details, contributing an extra BD-Rate gain of approximately 2.6\%. 
These results demonstrate both the effectiveness of each individual component and the synergy among them.  
To clearly demonstrate the effectiveness of the proposed Bi-FMT-based spatiotemporal alignment strategy, we visualize the feature errors relative to the target both before and after alignment. 
As illustrated in Figure~\ref{fig:aligned}, the unaligned reference features $F_{t-n}$ differ substantially from the current features, 
whereas the aligned features $F^{\text{aligned}}_{t-n}$ exhibit significantly reduced errors, with noticeably smaller high-error regions, 
thereby highlighting the effectiveness and synergistic contributions of the various components in our framework.

% The inclusion of the CTR module further decreases local feature errors, improving the precision of reconstructed features.

\begin{figure}[t]
\centering
\includegraphics[width=1\columnwidth]{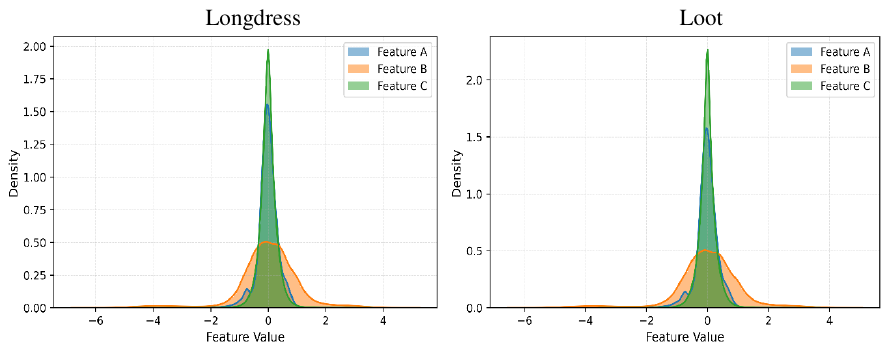} 
\caption{Residual distributions of predicted features on the \textit{Longdress} and \textit{Loot} sequences. 
Feature~A denotes the residual between the aligned feature after Bi-FMT and the ground-truth feature, 
Feature~C corresponds to the refined feature after CTR, 
and Feature~B represents the interpolated feature baseline.}
\label{fig:dist}
\end{figure}

\subsubsection{Residual Distribution Analysis} 
Figure~\ref{fig:dist} presents the residual distributions between the reconstructed and ground-truth features on the \textit{Longdress} and \textit{Loot} sequences. 
Specifically, Feature~A denotes the residual between the aligned feature after Bi-FMT and the ground-truth feature, 
Feature~B corresponds to the baseline obtained using simple nearest-neighbor interpolation, 
and Feature~C represents the refined feature produced by the CTR module at the decoder side. 
A distribution that is more concentrated around zero indicates smaller residuals and higher reconstruction fidelity. 
It can be observed that Feature~C exhibits the most compact distribution, suggesting that the CTR module effectively refines local misalignments in the decoded features. 
In comparison, Feature~A shows moderate dispersion, while Feature~B exhibits the widest spread due to the coarse nature of interpolation-based prediction. 
Overall, these results demonstrate the effectiveness of the proposed Bi-FMT in modeling non-rigid motion 
and highlight the capability of the CTR module in further enhancing the accuracy of locally reconstructed features.

\begin{figure}[t]
\centering
\includegraphics[width=0.9\columnwidth]{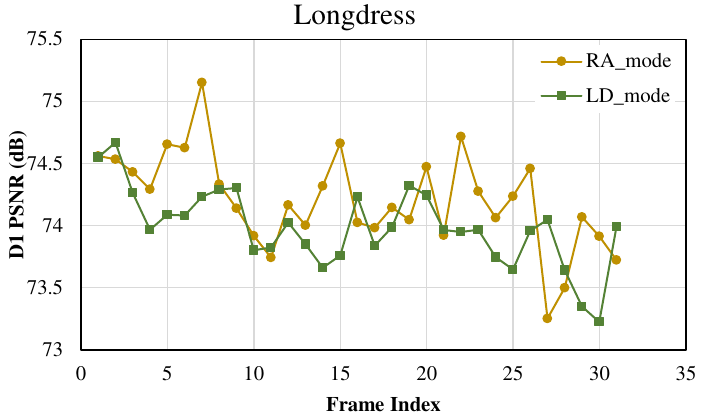} 
\caption{Quality fluctuation analysis on the Longdress sequence under both non-sequential (RA) and sequential (LD) coding modes.}
\label{fig:RA_wave}
\end{figure}

\subsubsection{Random Access (RA) Reference Strategy} 
Under the random access (RA) mode, dynamic point clouds are divided into fixed-length groups of frames (GOFs) and encoded following a predefined hierarchical structure, where higher-layer frames can reference both past and future frames for bidirectional prediction. Figure~\ref{fig:RA_wave} illustrates the quality fluctuations in two consecutive GOFs under both RA and sequential low-delay (LD) modes. Benefiting from bidirectional referencing, frames in the RA mode generally achieve noticeably higher quality compared to those in the LD mode, which relies solely on unidirectional prediction. However, the larger temporal gaps in RA may introduce less relevant or noisier features, causing a few frames (e.g., the 1st, 9th, and 16th) to exhibit slightly lower quality than their LD counterparts, which always reference the nearest past frame. Overall, the observed quality trends clearly demonstrate that the hierarchical RA strategy significantly improves dynamic point cloud compression by enhancing rate-distortion efficiency and enabling frame-level parallel encoding, making it suitable for real-time applications.
 
\section{Conclusion} 
This paper presents a novel deep compression framework for dynamic point clouds based on Bidirectional Feature-aligned Motion Transformation (Bi-FMT). The proposed Bi-FMT aligns features across both past and future frames to produce temporally consistent latent representations, which serve as predictive context within a conditional coding pipeline, forming a unified ``Motion + Conditional'' representation. Built upon this bidirectional feature alignment, a Cross-Transformer Refinement (CTR) module is incorporated at the decoder to adaptively refine locally aligned features, enhancing local consistency and restoring fine-grained spatial details. Furthermore, a Random Access (RA) reference strategy is introduced to enable frame-level parallel compression and eliminate sequential dependencies. Extensive experiments demonstrate that the proposed framework, integrating Bi-FMT and CTR, achieves state-of-the-art performance in both rate-distortion efficiency and computational complexity.

% \section*{Acknowledgments}
% This should be a simple paragraph before the References to thank those individuals and institutions who have supported your work on this article.

% {\appendix[Proof of the Zonklar Equations]
% Use $\backslash${\tt{appendix}} if you have a single appendix:
% Do not use $\backslash${\tt{section}} anymore after $\backslash${\tt{appendix}}, only $\backslash${\tt{section*}}.
% If you have multiple appendixes use $\backslash${\tt{appendices}} then use $\backslash${\tt{section}} to start each appendix.
% You must declare a $\backslash${\tt{section}} before using any $\backslash${\tt{subsection}} or using $\backslash${\tt{label}} ($\backslash${\tt{appendices}} by itself
%  starts a section numbered zero.)}

% {\appendices
% \section*{Proof of the First Zonklar Equation}
% \section*{Proof of the Second Zonklar Equation}
% Appendix two text goes here.}

% \section{References Section}

\bibliographystyle{IEEEtran}
\bibliography{egbib}

\begin{thebibliography}{10}
\providecommand{\url}[1]{#1}
\csname url@samestyle\endcsname
\providecommand{\newblock}{\relax}
\providecommand{\bibinfo}[2]{#2}
\providecommand{\BIBentrySTDinterwordspacing}{\spaceskip=0pt\relax}
\providecommand{\BIBentryALTinterwordstretchfactor}{4}
\providecommand{\BIBentryALTinterwordspacing}{\spaceskip=\fontdimen2\font plus
\BIBentryALTinterwordstretchfactor\fontdimen3\font minus \fontdimen4\font\relax}
\providecommand{\BIBforeignlanguage}[2]{{%
\expandafter\ifx\csname l@#1\endcsname\relax
\typeout{** WARNING: IEEEtran.bst: No hyphenation pattern has been}%
\typeout{** loaded for the language `#1'. Using the pattern for}%
\typeout{** the default language instead.}%
\else
\language=\csname l@#1\endcsname
\fi
#2}}
\providecommand{\BIBdecl}{\relax}
\BIBdecl

\bibitem{xu2018introduction}
Y.~Xu, K.~Zhang, L.~He, Z.~Jiang, and W.~Zhu, ``Introduction to point cloud compression,'' \emph{ZTE Communications}, vol.~16, no.~3, p.~3, 2018.

\bibitem{gao2023point}
P.~Gao, L.~Zhang, L.~Lei, and W.~Xiang, ``Point cloud compression based on joint optimization of graph transform and entropy coding for efficient data broadcasting,'' \emph{IEEE Transactions on Broadcasting}, vol.~69, no.~3, pp. 727--739, 2023.

\bibitem{li2019advanced}
L.~Li, Z.~Li, V.~Zakharchenko, J.~Chen, and H.~Li, ``Advanced 3d motion prediction for video-based dynamic point cloud compression,'' \emph{IEEE Transactions on Image Processing}, vol.~29, pp. 289--302, 2019.

\bibitem{xie2024roi}
L.~Xie, W.~Gao, H.~Zheng, and G.~Li, ``Roi-guided point cloud geometry compression towards human and machine vision,'' in \emph{Proceedings of the 32nd ACM International Conference on Multimedia}, 2024, pp. 3741--3750.

\bibitem{schwarz2018emerging}
S.~Schwarz, M.~Preda, V.~Baroncini, M.~Budagavi, P.~Cesar, P.~A. Chou, R.~A. Cohen, M.~Krivoku{\'c}a, S.~Lasserre, Z.~Li \emph{et~al.}, ``Emerging mpeg standards for point cloud compression,'' \emph{IEEE Journal on Emerging and Selected Topics in Circuits and Systems}, vol.~9, no.~1, pp. 133--148, 2018.

\bibitem{graziosi2020overview}
D.~Graziosi, O.~Nakagami, S.~Kuma, A.~Zaghetto, T.~Suzuki, and A.~Tabatabai, ``An overview of ongoing point cloud compression standardization activities: Video-based (v-pcc) and geometry-based (g-pcc),'' \emph{APSIPA Transactions on Signal and Information Processing}, vol.~9, p. e13, 2020.

\bibitem{li2024mpeg}
G.~Li, W.~Gao, and W.~Gao, ``Mpeg geometry-based point cloud compression (g-pcc) standard,'' in \emph{Point Cloud Compression: Technologies and Standardization}.\hskip 1em plus 0.5em minus 0.4em\relax Springer, 2024, pp. 135--165.

\bibitem{sullivan2012overview}
G.~J. Sullivan, J.-R. Ohm, W.-J. Han, and T.~Wiegand, ``Overview of the high efficiency video coding (hevc) standard,'' \emph{IEEE Transactions on circuits and systems for video technology}, vol.~22, no.~12, pp. 1649--1668, 2012.

\bibitem{santos2021block}
C.~Santos, M.~Gon{\c{c}}alves, G.~Corr{\^e}a, and M.~Porto, ``Block-based inter-frame prediction for dynamic point cloud compression,'' in \emph{2021 IEEE International Conference on Image Processing}.\hskip 1em plus 0.5em minus 0.4em\relax IEEE, 2021, pp. 3388--3392.

\bibitem{de2017motion}
R.~De and P.~A. Chou, ``Motion-compensated compression of dynamic voxelized point clouds,'' \emph{IEEE Transactions on Image Processing}, vol.~26, no.~8, pp. 3886--3895, 2017.

\bibitem{zhu2020view}
W.~Zhu, Z.~Ma, Y.~Xu, L.~Li, and Z.~Li, ``View-dependent dynamic point cloud compression,'' \emph{IEEE Transactions on Circuits and Systems for Video Technology}, vol.~31, no.~2, pp. 765--781, 2020.

\bibitem{cai2024distortion}
Z.~Cai, W.~Gao, G.~Li, and W.~Gao, ``Distortion propagation model-based v-pcc rate control for 3d point cloud broadcasting,'' \emph{IEEE Transactions on Broadcasting}, 2024.

\bibitem{huang2024temporal}
B.~Huang, D.~Lazzarotto, and T.~Ebrahimi, ``Temporal conditional coding for dynamic point cloud geometry compression,'' in \emph{ICASSP 2024-2024 IEEE International Conference on Acoustics, Speech and Signal Processing (ICASSP)}.\hskip 1em plus 0.5em minus 0.4em\relax IEEE, 2024, pp. 7920--7924.

\bibitem{zhang2024content}
J.~Zhang, J.~Zhang, W.~Ma, D.~Ding, and Z.~Ma, ``Content-aware rate control for geometry-based point cloud compression,'' \emph{IEEE Transactions on Circuits and Systems for Video Technology}, vol.~34, no.~10, pp. 9550--9561, 2024.

\bibitem{fan2022d}
T.~Fan, L.~Gao, Y.~Xu, Z.~Li, and D.~Wang, ``D-dpcc: Deep dynamic point cloud compression via 3d motion prediction,'' \emph{arXiv preprint arXiv:2205.01135}, 2022.

\bibitem{xia2023learning}
S.~Xia, T.~Fan, Y.~Xu, J.-N. Hwang, and Z.~Li, ``Learning dynamic point cloud compression via hierarchical inter-frame block matching,'' in \emph{Proceedings of the 31st ACM International Conference on Multimedia}, 2023, pp. 7993--8003.

\bibitem{jiang2023end}
Z.~Jiang, G.~Wang, G.~K. Tam, C.~Song, F.~W. Li, and B.~Yang, ``An end-to-end dynamic point cloud geometry compression in latent space,'' \emph{Displays}, vol.~80, p. 102528, 2023.

\bibitem{jiang2025mp}
Z.~Jiang, D.~Han, C.~Song, F.~Nan, and B.~Yang, ``Mp-dpcc: A motion proxy-based dynamic point cloud compression framework,'' in \emph{ICASSP 2025-2025 IEEE International Conference on Acoustics, Speech and Signal Processing (ICASSP)}.\hskip 1em plus 0.5em minus 0.4em\relax IEEE, 2025, pp. 1--5.

\bibitem{akhtar2024inter}
A.~Akhtar, Z.~Li, and G.~Van~der Auwera, ``Inter-frame compression for dynamic point cloud geometry coding,'' \emph{IEEE Transactions on Image Processing}, vol.~33, pp. 584--594, 2024.

\bibitem{liu2024encoding}
G.~Liu, J.~Zhu, D.~Ding, and Z.~Ma, ``Encoding auxiliary information to restore compressed point cloud geometry,'' in \emph{Proceedings of the Thirty-Third International Joint Conference on Artificial Intelligence, IJCAI-24}, 2024, pp. 2189--2197.

\bibitem{zhang2025adadpcc}
C.~Zhang and W.~Gao, ``Adadpcc: Adaptive rate control and rate-distortion-complexity optimization for dynamic point cloud compression,'' in \emph{Proceedings of the AAAI Conference on Artificial Intelligence}, vol.~39, no.~12, 2025, pp. 13\,188--13\,196.

\bibitem{2025A}
J.~Wang, R.~Xue, J.~Li, D.~Ding, Y.~Lin, and Z.~Ma, ``A versatile point cloud compressor using universal multiscale conditional coding – part i: Geometry,'' \emph{Pattern Analysis and Machine Intelligence, IEEE Transactions on}, vol.~47, no.~1, pp. 252--268, 2025.

\bibitem{chen2023patchvvc}
R.~Chen, M.~Xiao, D.~Yu, G.~Zhang, and Y.~Liu, ``Patchvvc: A real-time compression framework for streaming volumetric videos,'' in \emph{Proceedings of the 14th Conference on ACM Multimedia Systems}, 2023, pp. 119--129.

\bibitem{mekuria2016design}
R.~Mekuria, K.~Blom, and P.~Cesar, ``Design, implementation, and evaluation of a point cloud codec for tele-immersive video,'' \emph{IEEE Transactions on Circuits and Systems for Video Technology}, vol.~27, no.~4, pp. 828--842, 2016.

\bibitem{wang2022sparse}
J.~Wang, D.~Ding, Z.~Li, X.~Feng, C.~Cao, and Z.~Ma, ``Sparse tensor-based multiscale representation for point cloud geometry compression,'' \emph{IEEE Transactions on Pattern Analysis and Machine Intelligence}, vol.~45, no.~7, pp. 9055--9071, 2022.

\bibitem{zhang2023yoga}
J.~Zhang, T.~Chen, D.~Ding, and Z.~Ma, ``Yoga: Yet another geometry-based point cloud compressor,'' in \emph{Proceedings of the 31st ACM International Conference on Multimedia}, 2023, pp. 9070--9081.

\bibitem{wang2024versatile}
J.~Wang, R.~Xue, J.~Li, D.~Ding, Y.~Lin, and Z.~Ma, ``A versatile point cloud compressor using universal multiscale conditional coding--part i: Geometry,'' \emph{IEEE transactions on pattern analysis and machine intelligence}, 2024.

\bibitem{wang2021multiscale}
J.~Wang, D.~Ding, Z.~Li, and Z.~Ma, ``Multiscale point cloud geometry compression,'' in \emph{2021 Data Compression Conference}.\hskip 1em plus 0.5em minus 0.4em\relax IEEE, 2021, pp. 73--82.

\bibitem{fu2022octattention}
C.~Fu, G.~Li, R.~Song, W.~Gao, and S.~Liu, ``Octattention: Octree-based large-scale contexts model for point cloud compression,'' in \emph{Proceedings of the AAAI conference on artificial intelligence}, vol.~36, no.~1, 2022, pp. 625--633.

\bibitem{cui2023octformer}
M.~Cui, J.~Long, M.~Feng, B.~Li, and H.~Kai, ``Octformer: Efficient octree-based transformer for point cloud compression with local enhancement,'' in \emph{Proceedings of the AAAI Conference on Artificial Intelligence}, vol.~37, no.~1, 2023, pp. 470--478.

\bibitem{nguyen2023lossless}
D.~T. Nguyen and A.~Kaup, ``Lossless point cloud geometry and attribute compression using a learned conditional probability model,'' \emph{IEEE Transactions on Circuits and Systems for Video Technology}, vol.~33, no.~8, pp. 4337--4348, 2023.

\bibitem{guo2024tsc}
Z.~Guo, Y.~Zhang, L.~Zhu, H.~Wang, and G.~Jiang, ``Tsc-pcac: Voxel transformer and sparse convolution-based point cloud attribute compression for 3d broadcasting,'' \emph{IEEE Transactions on Broadcasting}, 2024.

\bibitem{liu2019flownet3d}
X.~Liu, C.~R. Qi, and L.~J. Guibas, ``Flownet3d: Learning scene flow in 3d point clouds,'' in \emph{Proceedings of the IEEE/CVF conference on computer vision and pattern recognition}, 2019, pp. 529--537.

\bibitem{qi2017pointnet++}
C.~R. Qi, L.~Yi, H.~Su, and L.~J. Guibas, ``Pointnet++: Deep hierarchical feature learning on point sets in a metric space,'' \emph{Advances in neural information processing systems}, vol.~30, 2017.

\bibitem{wu2020pointpwc}
W.~Wu, Z.~Y. Wang, Z.~Li, W.~Liu, and L.~Fuxin, ``Pointpwc-net: Cost volume on point clouds for (self-) supervised scene flow estimation,'' in \emph{Computer Vision--ECCV 2020: 16th European Conference, Glasgow, UK, August 23--28, 2020, Proceedings, Part V 16}.\hskip 1em plus 0.5em minus 0.4em\relax Springer, 2020, pp. 88--107.

\bibitem{wei2021pv}
Y.~Wei, Z.~Wang, Y.~Rao, J.~Lu, and J.~Zhou, ``Pv-raft: Point-voxel correlation fields for scene flow estimation of point clouds,'' in \emph{Proceedings of the IEEE/CVF conference on computer vision and pattern recognition}, 2021, pp. 6954--6963.

\bibitem{wu2019pointconv}
W.~Wu, Z.~Qi, and L.~Fuxin, ``Pointconv: Deep convolutional networks on 3d point clouds,'' in \emph{Proceedings of the IEEE/CVF Conference on computer vision and pattern recognition}, 2019, pp. 9621--9630.

\bibitem{hu2021fvc}
Z.~Hu, G.~Lu, and D.~Xu, ``Fvc: A new framework towards deep video compression in feature space,'' in \emph{Proceedings of the IEEE/CVF conference on computer vision and pattern recognition}, 2021, pp. 1502--1511.

\bibitem{hu2022coarse}
Z.~Hu, G.~Lu, J.~Guo, S.~Liu, W.~Jiang, and D.~Xu, ``Coarse-to-fine deep video coding with hyperprior-guided mode prediction,'' in \emph{Proceedings of the IEEE/CVF Conference on Computer Vision and Pattern Recognition}, 2022, pp. 5921--5930.

\bibitem{lu2019dvc}
G.~Lu, W.~Ouyang, D.~Xu, X.~Zhang, C.~Cai, and Z.~Gao, ``Dvc: An end-to-end deep video compression framework,'' in \emph{Proceedings of the IEEE/CVF conference on computer vision and pattern recognition}, 2019, pp. 11\,006--11\,015.

\bibitem{li2024neural}
J.~Li, B.~Li, and Y.~Lu, ``Neural video compression with feature modulation,'' in \emph{Proceedings of the IEEE/CVF Conference on Computer Vision and Pattern Recognition}, 2024, pp. 26\,099--26\,108.

\bibitem{jia2025towards}
Z.~Jia, B.~Li, J.~Li, W.~Xie, L.~Qi, H.~Li, and Y.~Lu, ``Towards practical real-time neural video compression,'' in \emph{Proceedings of the Computer Vision and Pattern Recognition Conference}, 2025, pp. 12\,543--12\,552.

\bibitem{choy20194d}
C.~Choy, J.~Gwak, and S.~Savarese, ``4d spatio-temporal convnets: Minkowski convolutional neural networks,'' in \emph{Proceedings of the IEEE Conference on Computer Vision and Pattern Recognition}, 2019, pp. 3075--3084.

\bibitem{vaswani2017attention}
A.~Vaswani, N.~Shazeer, N.~Parmar, J.~Uszkoreit, L.~Jones, A.~N. Gomez, {\L}.~Kaiser, and I.~Polosukhin, ``Attention is all you need,'' \emph{Advances in neural information processing systems}, vol.~30, 2017.

\bibitem{zhao2020exploring}
H.~Zhao, J.~Jia, and V.~Koltun, ``Exploring self-attention for image recognition,'' in \emph{Proceedings of the IEEE/CVF conference on computer vision and pattern recognition}, 2020, pp. 10\,076--10\,085.

\bibitem{zhao2021point}
H.~Zhao, L.~Jiang, J.~Jia, P.~H. Torr, and V.~Koltun, ``Point transformer,'' in \emph{Proceedings of the IEEE/CVF international conference on computer vision}, 2021, pp. 16\,259--16\,268.

\bibitem{li2021deep}
J.~Li, B.~Li, and Y.~Lu, ``Deep contextual video compression,'' \emph{Advances in Neural Information Processing Systems}, vol.~34, pp. 18\,114--18\,125, 2021.

\bibitem{shannon1948mathematical}
C.~E. Shannon, ``A mathematical theory of communication,'' \emph{The Bell system technical journal}, vol.~27, no.~3, pp. 379--423, 1948.

\bibitem{balle2018variational}
J.~Ball{\'e}, D.~Minnen, S.~Singh, S.~J. Hwang, and N.~Johnston, ``Variational image compression with a scale hyperprior,'' \emph{arXiv preprint arXiv:1802.01436}, 2018.

\bibitem{minnen2018joint}
D.~Minnen, J.~Ball{\'e}, and G.~D. Toderici, ``Joint autoregressive and hierarchical priors for learned image compression,'' \emph{Advances in neural information processing systems}, vol.~31, 2018.

\bibitem{kamisli2024dcc_vbrlic}
F.~Kamisli, F.~Racap{\'e}, and H.~Choi, ``Variable-rate learned image compression with multi-objective optimization and quantization-reconstruction offsets,'' 2024.

\bibitem{bross2021overview}
B.~Bross, Y.-K. Wang, Y.~Ye, S.~Liu, J.~Chen, G.~J. Sullivan, and J.-R. Ohm, ``Overview of the versatile video coding (vvc) standard and its applications,'' \emph{IEEE Transactions on Circuits and Systems for Video Technology}, vol.~31, no.~10, pp. 3736--3764, 2021.

\bibitem{owlii}
Y.~L. Yi~Xu and Z.~Wen, ``Owlii dynamic human mesh sequence dataset,'' \emph{ISO/IEC JTC1/SC29 Joint WG11/WG1 (MPEG/JPEG) input document WG11M40059/WG1M74006}, vol.~7, no.~8, p.~11, 2017.

\bibitem{mpeg2022}
\emph{ISO/IEC 23090-12:2021 Information technology — Coded representation of immersive media — Part 12: MPEG Point Cloud Compression (MPEG-PCC)}, ISO/IEC JTC1/SC29 Std., 2021.

\bibitem{d20178i}
E.~d’Eon, B.~Harrison, T.~Myers, and P.~A. Chou, ``8i voxelized full bodies-a voxelized point cloud dataset,'' \emph{ISO/IEC JTC1/SC29 Joint WG11/WG1 (MPEG/JPEG) input document WG11M40059/WG1M74006}, vol.~7, no.~8, p.~11, 2017.

\bibitem{kinga2015method}
D.~Kinga, J.~B. Adam \emph{et~al.}, ``A method for stochastic optimization,'' in \emph{International conference on learning representations (ICLR)}, vol.~5, no.~6.\hskip 1em plus 0.5em minus 0.4em\relax California;, 2015.

\bibitem{pan2024patchdpcc}
Z.~Pan, M.~Xiao, X.~Han, D.~Yu, G.~Zhang, and Y.~Liu, ``Patchdpcc: A patchwise deep compression framework for dynamic point clouds,'' in \emph{Proceedings of the AAAI Conference on Artificial Intelligence}, vol.~38, no.~5, 2024, pp. 4406--4414.

\end{thebibliography}

\newpage

 

% \end{IEEEbiography}



\vfill

\end{document}